%% file: paper.tex
\documentclass[]{fairmeta}
\usepackage{amsthm,amsmath,amssymb}
\usepackage{graphicx}
\usepackage{natbib}
\usepackage{subfig}
\usepackage{subcaption}
\usepackage{booktabs}
\usepackage{tabularx}
\usepackage{array}
\usepackage{colortbl}
\usepackage{algorithm}
\usepackage{algorithmic}
\usepackage{tikz}
\usetikzlibrary{arrows.meta,positioning,fit,calc,shapes.geometric,shapes.symbols,shapes.misc,backgrounds,decorations.pathmorphing,patterns}
\newtheorem{theorem}{Theorem}[]

\newtheorem{remark1}[theorem]{Remark}

\definecolor{metawarm}{HTML}{B85C38}
\definecolor{figblue}{HTML}{2F6FAB}
\definecolor{figwarm}{HTML}{C55A2A}
\definecolor{figgreen}{HTML}{2F855A}
\definecolor{figpurple}{HTML}{6B5BB8}
\definecolor{figgray}{HTML}{5B6470}
\definecolor{figbg}{HTML}{F6F8FB}
\definecolor{figwarmbg}{HTML}{FFF3EA}
\definecolor{figgreenbg}{HTML}{EEF8F1}
\definecolor{figpurplebg}{HTML}{F3F1FC}
\definecolor{figgraybg}{HTML}{F3F4F6}

\tikzset{
  figlabel/.style={font=\sffamily\scriptsize, align=center},
  figtitle/.style={font=\sffamily\bfseries\scriptsize, align=center},
  figmodel/.style={figlabel, draw=figblue!85!black, fill=figblue!8, rounded corners=5pt, line width=0.75pt},
  figwam/.style={figlabel, draw=figwarm!90!black, fill=figwarmbg, rounded corners=5pt, line width=0.85pt},
  figharness/.style={figlabel, draw=figpurple!85!black, fill=figpurplebg, rounded corners=5pt, line width=0.85pt},
  figtool/.style={figlabel, draw=figgray!80!black, fill=white, rounded corners=4pt, line width=0.65pt},
  figdata/.style={figlabel, draw=figgreen!80!black, fill=figgreenbg, rounded corners=4pt, line width=0.75pt},
  figeval/.style={figlabel, draw=figwarm!90!black, fill=figwarm!9, rounded corners=4pt, line width=0.75pt},
  figcontract/.style={figlabel, draw=figgray!75!black, fill=figgraybg, rounded corners=4pt, line width=0.75pt},
  figarrow/.style={-{Latex[length=2mm]}, line width=0.85pt, draw=figblue!85!black},
  figpred/.style={-{Latex[length=2mm]}, line width=0.85pt, draw=figwarm!90!black},
  figgate/.style={-{Latex[length=2mm]}, line width=0.85pt, draw=figwarm!90!black, dashed},
  figlearn/.style={-{Latex[length=2mm]}, line width=0.85pt, dashed, draw=figgreen!85!black},
  figstd/.style={line width=0.7pt, draw=figgray!75!black},
  figcompat/.style={line width=0.8pt, draw=figgray!75!black},
  figdependency/.style={line width=0.75pt, dashed, draw=figgray!75!black},
  figprobe/.style={-{Latex[length=2mm]}, line width=0.8pt, dashed, draw=figwarm!90!black},
  figdivider/.style={line width=0.55pt, draw=black!70},
}
\input{chapters/00_figure_styles}

\newcolumntype{P}[1]{>{\raggedright\arraybackslash}p{#1}}
\newcolumntype{Y}{>{\raggedright\arraybackslash}X}

\newcommand{\existingbadge}{\textcolor{figblue!75!black}{\sffamily\bfseries Existing}}
\newcommand{\proposedbadge}{\textcolor{figpurple!75!black}{\sffamily\bfseries Proposed}}
\newcommand{\bothbadge}{\textcolor{figgreen!60!black}{\sffamily\bfseries Both}}

\newcommand{\vla}{Vision-Language-Action}

\newtcolorbox{contributorslot}[1]{
  colback=metabg,
  colframe=metablue!55!black,
  title={Contributor slot: #1},
  fonttitle=\sffamily\bfseries,
  breakable,
  arc=4pt
}
\usepackage{graphicx}

\title{%
\begin{minipage}{\textwidth}
\raggedright
\mbox{From World Action Models to}\\[-0.75em]
\mbox{Embodied Brains: A Roadmap for}\\[-0.75em]
Open-World Physical Intelligence
\end{minipage}%
}

\author{Yuanzhi Liang, Xufeng Zhan, Haibin Huang, Chi Zhang, Xuelong Li}
\metadata[Corresponding Author]{Xuelong Li(\email{xuelong$\_$li@ieee.org})}

\begin{document}

\abstract{
\input{chapters/00_abstract}
}

\maketitle

\vspace{-0.1em}

\input{chapters/01_introduction}
\input{chapters/01a_visual_overview}
\input{chapters/02_evolution_toward_wams}
\input{chapters/03_current_barriers}
\input{chapters/04_coevolution_roadmap}
\input{chapters/05_conclusion}



\bibliography{paper}
\bibliographystyle{authordate1}

\end{document}

%% file: chapters/00_figure_styles.tex
\tikzset{
  vividpanel/.style={figlabel, draw=figgray!70, fill=figbg, rounded corners=9pt, line width=0.9pt, inner sep=6pt},
  vividpanelblue/.style={vividpanel, draw=figblue!70!black, fill=figblue!7},
  vividpanelwarm/.style={vividpanel, draw=figwarm!80!black, fill=figwarmbg},
  vividpanelgreen/.style={vividpanel, draw=figgreen!70!black, fill=figgreenbg},
  vividpanelpurple/.style={vividpanel, draw=figpurple!70!black, fill=figpurplebg},
  vividbadge/.style={font=\sffamily\bfseries\scriptsize, rounded corners=3pt, inner xsep=3.5pt, inner ysep=1.6pt, draw=figgray!65, fill=white, align=center},
  vividcallout/.style={figlabel, draw=figgray!65, fill=white, rounded corners=6pt, line width=0.75pt, inner sep=4pt},
  vividicon/.style={circle, draw=figgray!75, fill=white, line width=0.75pt, minimum size=0.62cm, inner sep=0pt},
  vividcard/.style={figlabel, draw=figgray!65, fill=white, rounded corners=5pt, line width=0.75pt, minimum width=1.75cm, minimum height=0.72cm},
  vividstage/.style={figtitle, draw=figgray!70, fill=white, rounded corners=6pt, line width=0.9pt, minimum width=2.25cm, minimum height=0.9cm},
  sketcharrow/.style={-{Latex[length=2mm]}, line width=1.0pt, draw=figgray!85, decorate, decoration={random steps, segment length=8pt, amplitude=0.35pt}},
  sketchblue/.style={-{Latex[length=2mm]}, line width=1.0pt, draw=figblue!85!black, decorate, decoration={random steps, segment length=8pt, amplitude=0.35pt}},
  sketchpred/.style={-{Latex[length=2mm]}, line width=1.0pt, draw=figwarm!90!black, decorate, decoration={random steps, segment length=8pt, amplitude=0.35pt}},
  sketchlearn/.style={-{Latex[length=2mm]}, line width=1.0pt, dashed, draw=figgreen!85!black},
  sketchstd/.style={line width=0.9pt, draw=figgray!70, dashed},
  figcontractlink/.style={line width=0.75pt, draw=figgray!75!black, dashed},
  candidatepanel/.style={draw=figgray!60, fill=figbg, rounded corners=6pt, line width=0.75pt, inner sep=5pt},
  candidateheading/.style={font=\sffamily\bfseries\scriptsize, text=figgray!90!black, align=center},
  candidatebox/.style={font=\sffamily\scriptsize, align=center, rounded corners=4pt, line width=0.75pt, inner sep=3pt},
  candidateannotation/.style={font=\sffamily\tiny, align=center, text=figgray!90!black}
}

\tikzset{
  pics/vcamera/.style={code={
    \draw[fill=figblue!10, draw=figblue!80!black, rounded corners=1pt, line width=.7pt] (-.30,-.18) rectangle (.30,.18);
    \draw[fill=white, draw=figblue!80!black, line width=.6pt] (0,0) circle (.10);
    \draw[fill=figblue!18, draw=figblue!80!black, rounded corners=.5pt, line width=.6pt] (-.20,.18) rectangle (.05,.30);
    \draw[draw=figblue!80!black, line width=.6pt] (.19,.08)--(.30,.16);
  }},
  pics/vrobot/.style={code={
    \draw[fill=figgreen!13, draw=figgreen!70!black, rounded corners=1pt, line width=.7pt] (-.30,-.30) rectangle (.30,-.12);
    \draw[draw=figgreen!70!black, line width=1.2pt] (-.10,-.12)--(.05,.08)--(.27,.18);
    \draw[fill=white, draw=figgreen!70!black, line width=.7pt] (-.10,-.12) circle (.055);
    \draw[fill=white, draw=figgreen!70!black, line width=.7pt] (.05,.08) circle (.055);
    \draw[fill=white, draw=figgreen!70!black, line width=.7pt] (.27,.18) circle (.055);
    \draw[draw=figgreen!70!black, line width=.8pt] (.31,.20)--(.42,.25);
    \draw[draw=figgreen!70!black, line width=.8pt] (.31,.16)--(.42,.11);
  }},
  pics/vbrain/.style={code={
    \draw[fill=figwarm!13, draw=figwarm!90!black, line width=.75pt]
      (-.28,-.04) .. controls (-.36,.16) and (-.18,.32) .. (.02,.26)
      .. controls (.18,.38) and (.38,.18) .. (.28,.02)
      .. controls (.42,-.18) and (.18,-.32) .. (.00,-.24)
      .. controls (-.14,-.34) and (-.36,-.22) .. (-.28,-.04);
    \draw[draw=figwarm!90!black, line width=.55pt] (-.15,.12) .. controls (.02,.04) and (.06,.22) .. (.18,.10);
    \draw[draw=figwarm!90!black, line width=.55pt] (-.16,-.06) .. controls (.02,-.12) and (.08,.02) .. (.20,-.08);
  }},
  pics/vwam/.style={code={
    \draw[fill=figwarm!12, draw=figwarm!90!black, line width=.75pt] (0,.34)--(.30,.02)--(.12,-.34)--(-.18,-.30)--(-.32,.02)--cycle;
    \draw[draw=figwarm!90!black, line width=.55pt] (0,.34)--(.03,-.22);
    \draw[draw=figwarm!90!black, line width=.55pt] (-.32,.02)--(.30,.02);
    \draw[draw=figwarm!90!black, line width=.55pt] (-.18,-.30)--(.10,.02);
  }},
  pics/vharness/.style={code={
    \draw[fill=figpurple!10, draw=figpurple!80!black, rounded corners=1pt, line width=.75pt] (-.34,-.25) rectangle (.34,.25);
    \draw[draw=figpurple!80!black, line width=.55pt] (-.24,.11)--(.20,.11);
    \draw[draw=figpurple!80!black, line width=.55pt] (-.24,0)--(.10,0);
    \draw[draw=figpurple!80!black, line width=.55pt] (-.24,-.11)--(.24,-.11);
    \draw[fill=figpurple!70, draw=figpurple!80!black, line width=.45pt] (.23,.11) circle (.025);
    \draw[fill=figgreen!70, draw=figgreen!80!black, line width=.45pt] (.14,0) circle (.025);
  }},
  pics/vtool/.style={code={
    \draw[fill=figgray!12, draw=figgray!85, rounded corners=1pt, line width=.7pt] (-.32,-.20) rectangle (.32,.20);
    \draw[draw=figgray!85, line width=.7pt] (-.12,.20) arc (180:0:.12 and .08);
    \draw[draw=figgray!85, line width=.75pt] (-.22,-.04)--(.22,-.04);
    \draw[draw=figgray!85, line width=.75pt] (-.06,.05)--(.06,-.13);
  }},
  pics/vshield/.style={code={
    \draw[fill=figgreen!12, draw=figgreen!75!black, line width=.75pt] (0,.32)--(.26,.20)--(.20,-.14)--(0,-.34)--(-.20,-.14)--(-.26,.20)--cycle;
    \draw[draw=figgreen!75!black, line width=.75pt] (-.12,.00)--(-.03,-.10)--(.15,.12);
  }},
  pics/vtrace/.style={code={
    \draw[fill=white, draw=figblue!80!black, rounded corners=1pt, line width=.7pt] (-.26,-.32) rectangle (.26,.32);
    \draw[fill=figblue!15, draw=figblue!80!black, line width=.45pt] (.08,.32)--(.26,.14)--(.08,.14)--cycle;
    \draw[draw=figblue!80!black, line width=.45pt] (-.16,.12)--(.10,.12);
    \draw[draw=figblue!80!black, line width=.45pt] (-.16,.00)--(.16,.00);
    \draw[draw=figblue!80!black, line width=.45pt] (-.16,-.12)--(.12,-.12);
  }},
  pics/vworld/.style={code={
    \draw[fill=figgreen!9, draw=figgreen!75!black, rounded corners=1pt, line width=.7pt] (-.36,-.28) rectangle (.36,.28);
    \draw[fill=figwarm!55, draw=figwarm!85!black, line width=.45pt] (.20,.12) circle (.055);
    \draw[draw=figgreen!75!black, line width=.55pt] (-.30,-.10) .. controls (-.10,.05) and (.02,-.20) .. (.30,.02);
    \draw[draw=figgreen!75!black, line width=.55pt] (-.32,-.20)--(.32,-.20);
  }},
  pics/vgate/.style={code={
    \draw[fill=figpurple!9, draw=figpurple!75!black, rounded corners=1pt, line width=.7pt] (-.30,-.22) rectangle (.30,.22);
    \draw[draw=figpurple!75!black, line width=.75pt] (-.16,-.01)--(-.05,-.12)--(.17,.12);
  }},
  pics/vloop/.style={code={
    \draw[-{Latex[length=1.4mm]}, draw=figgreen!80!black, line width=.75pt] (155:.24) arc (155:-80:.24);
    \draw[-{Latex[length=1.4mm]}, draw=figgreen!80!black, line width=.75pt] (-25:.24) arc (-25:200:.24);
  }}
}

%% file: chapters/00_abstract.tex
Artificial general intelligence ultimately requires agents that can learn, reason, and act in the physical world. Action models, vision-language-action policies, and world models have made this ambition concrete. World Action Models (WAMs) are especially promising because they connect candidate interventions to predicted physical consequences, supporting consequence-aware decisions. Yet progress does not accumulate readily. Many action-centric systems bind outputs to action spaces and controllers. Predictive systems expose different variables; datasets, tasks, and runtimes use incompatible conventions. Scaling parameters or trajectories alone does not resolve these interfaces. We survey the evolution toward WAMs and organize these limitations into three coupled gaps in model roles and representations, objectives and standardization, and system composition. We then propose a co-evolution roadmap for scalable physical intelligence. At its center is the \emph{embodied brain}, a long-term model target for physical reasoning. It integrates multimodal context, compares possible interventions, and communicates an intended state transition or capability request instead of actuator commands. Current WAMs provide promising prototypes for the predictive capabilities this model may require without fixing its final architecture. A physical harness grounds the brain output, resolves tools and controllers, verifies execution, and records the resulting trace. Shared contracts make heterogeneous data, tasks, and components comparable while preserving embodiment details. Closed-loop post-training turns verified interaction into reusable experience. By separating general physical reasoning from local execution, the roadmap makes components independently testable, mutually compatible, and jointly improvable. The resulting physical-intelligence stack offers an actionable path toward adaptive, self-improving agents and more general intelligence grounded in real-world interaction.

%% file: chapters/01_introduction.tex
\section{Introduction}

A long-standing ambition of artificial intelligence is to build general-purpose agents that can perceive, reason, act, and improve across open-ended tasks. Recent progress in large language models and multimodal foundation models has made this ambition increasingly concrete in digital domains. Models can now summarize knowledge, write code, use tools, follow instructions, and participate in multi-step workflows. Yet a purely digital notion of intelligence remains incomplete. Human intelligence is not only linguistic or symbolic; it is grounded in space, time, objects, forces, affordances, and consequences. A system that aspires to general-purpose agency must eventually move beyond predicting text or images and acquire the ability to understand and intervene in the physical world. In this sense, physical intelligence is not a peripheral application of AGI, but a central test of whether intelligent systems can connect perception, prediction, decision making, and action under real-world constraints.

This view is consistent with several influential perspectives on the future of AI. The ``bitter lesson'' emphasizes general methods that continue to benefit from computation and data \citep{sutton2019bitterlesson}. Recent discussions of an ``era of experience'' further argue that future agents should learn not only from static human-generated corpora, but also from their own interaction traces and consequences \citep{silver2025eraexperience}. In parallel, world-model and spatial-intelligence perspectives argue that intelligent behavior requires internal representations of the external world that support prediction, planning, and control \citep{lecun2022path,li2025spatialintelligence}. Together, these perspectives point to a common requirement: the next stage of AI should not only learn from observations, but also operate in closed loops of perception, intervention, feedback, and self-improvement.

Embodied AI and robot learning make this requirement concrete. In a physical environment, an action changes the world, and successful behavior often depends on anticipating that change. Recent systems span language-grounded planning, multimodal goal specification, and \vla{} policies that map visual, linguistic, and proprioceptive context to robot actions \citep{ahn2022saycan,driess2023palme,jiang2022vima,brohan2022rt1,brohan2023rt2,kim2024openvla,octo2024,black2024pi0}. These systems broaden task and embodiment coverage, but observation-to-action prediction is only one part of physical agency. An embodied agent must also maintain task-relevant state, compare possible interventions, estimate feasibility and uncertainty, and delegate execution when specialized control is required. This reflects a classical insight from motor control and affordance theory: perception supports behavior when it is organized around action and consequence rather than passive description \citep{wolpert1998multiple,gibson1979ecological}.

The action-oriented interface also exposes a scaling problem. Physical commands are defined relative to a robot's morphology, coordinate frames, control mode, and sensing configuration. Open X-Embodiment coarsely aligns heterogeneous actions into a common end-effector format while retaining dataset-dependent normalization and control conventions \citep{oneill2023openx}. Octo accommodates new observations and action spaces through configurable inputs, output heads, and downstream fine-tuning \citep{octo2024}. These are important advances in cross-embodiment learning. They also show that reuse depends on an explicit mapping between shared model representations and local control. When this mapping remains implicit inside a policy, changing a gripper, controller, or morphology can alter both the execution interface and the learned prediction target.

Informed by the emerging literature, we use \emph{World Action Model} (WAM) to denote an action-conditioned predictive model that makes intervention outcomes explicit for planning, verification, control, or learning \citep{wang2026wam}. Under this definition, a WAM estimates what may happen if a candidate action, skill, or tool call is executed. Existing systems instantiate this capability through joint image--action prediction, unified video--action modeling, 3D scene prediction, world--language--action modeling, and structured world--action stacks \citep{wu2023gr1,guo2024predictionactionvisualpolicy,li2025unifiedvideoactionmodel,lu2025gwm,yang2026wla,kairos2026}. This diversity demonstrates the promise of consequence modeling. It also leaves open what a WAM should predict, how that information should be represented, and how another system should consume it.

These observations reveal a broader scaling problem. Physical intelligence must scale in at least three complementary senses. Models must acquire stronger spatial and physical representations; learned capabilities must transfer across tasks, tools, and embodiments; and advances from different datasets and systems must accumulate. Increasing parameters or trajectories primarily addresses the first requirement. The latter two also require interfaces that preserve the meaning of predictions, actions, and outcomes across heterogeneous systems. Diversity is not itself the obstacle. The obstacle is diversity whose semantics remain hidden.

We organize this problem into three coupled challenges. The first is a \emph{model and representation gap}. Current systems combine language and vision-language models, action tokenizers, diffusion policies, video generators, and multimodal backbones. These components strengthen semantic grounding or generative capacity, but they do not automatically provide a persistent and actionable account of physical state. Spatial structure, contact-sensitive reasoning, calibrated uncertainty, and recovery remain open challenges; recent evaluation reports failures in spatial reasoning, contact prediction, and physical consistency \citep{robowmbench2026}. The central model role is also unsettled. The field has not converged on which physical capabilities it should learn, which consequences it should expose, or which responsibilities should remain downstream.

The second challenge is an \emph{objective and standardization gap}. End-task success cannot by itself reveal whether a system predicted the right consequence, exploited a narrow benchmark regularity, or relied on a strong controller. Existing evaluations address different layers, including executability, embodied reasoning, benchmark construction, and safety \citep{robowmbench2026,luo2025robobench,zhang2026a2eval,cui2026liberosafety,yin2026roboshackles}. Data collection, task definitions, supervision, and evaluation protocols also differ across projects. Without shared semantics for interventions, predicted changes, and outcomes, these resources cannot provide fully comparable supervision. More data then increases volume without necessarily increasing reusable physical experience.

The third challenge is an \emph{ecosystem and systems gap}. A single model need not own general reasoning, low-level control, simulation, verification, memory, safety, and every execution service. Digital agents separate these roles through harnesses that manage tool calls, state, guardrails, tracing, and multi-step execution \citep{openai2025agentsguide,openai2025agentssdk,anthropic2024buildingeffectiveagents}. Language, code, and structured APIs provide their communication medium. Physical agents lack an equally mature interface for spatial relations, coordinate frames, intended state transitions, uncertainty, and controller constraints. Consequently, each project often reconnects perception, reasoning, and control through a system-specific runtime, limiting component substitution and trace reuse.

The roadmap addresses these gaps through explicit model and system contracts. For current WAM research, we propose a prediction contract rather than a mandatory output modality. Given physical context and a candidate intervention, the model should expose a decision-relevant consequence together with its horizon, reference frame, uncertainty, and validity conditions. The consequence may be represented through video, geometry, structured state, predictive latents, or a hybrid. A model may also predict actions, but an action vector alone does not make the expected consequence explicit.

At the center of the systems roadmap, we define the \emph{embodied brain} as a long-term model target. It integrates physical observations and task context, reasons over possible interventions, and emits an intermediate intent representation. This representation communicates an intended state transition or capability request rather than embodiment-specific actuator commands. The embodied brain is distinct from the complete \emph{physical-intelligence stack}, which also contains a physical harness, tool models, controllers, verifiers, sensors, bodies, and environments. Current WAMs are promising prototypes for predictive capabilities that this model may require; they are neither equivalent to the embodied brain nor required to remain a separate module.

This separation is intended to make physical reasoning reusable. A change of gripper, controller, or morphology should primarily alter the declared capability and its adapter instead of forcing all model-level reasoning to be learned again. This is a design hypothesis, not a guaranteed consequence of modularity. It should be tested through representation grounding, tool substitution, and cross-embodiment adaptation. Joint or end-to-end training remains compatible with the roadmap when the intermediate responsibilities and transformations remain inspectable.

Existing systems address complementary parts of this view. The original RoboBrain is a large-scale knowledge engine, while recent RoboBrain work proposes a unified manipulation model \citep{saxena2014robobrain,ji2025robobrain}. RoboOS defines a hierarchical embodied framework, and Embodied.cpp focuses on portable inference across heterogeneous robots \citep{tan2025roboos,xu2026embodiedcpp}. We do not treat these systems as equivalent. Their different objectives instead illustrate why a shared model role and explicit interfaces are still needed.

Accordingly, this paper treats WAM research as a starting point rather than a self-contained endpoint. The co-evolution roadmap connects the embodied brain to a physical harness, tool capabilities, data and task contracts, and closed-loop learning. Physical interaction is recorded as a replayable trace that preserves the decision context, translation process, execution state, verifier outputs, and outcome. Such traces can train and diagnose WAM prototypes, future brain models, harness adapters, and tool models. Evaluation then determines which traces are reliable enough to return to post-training.

This perspective also changes how WAMs should be evaluated. Pixel reconstruction can reward appearance factors that are weakly coupled to action. Final task success can hide whether performance came from physical foresight, reactive robustness, controller quality, or benchmark structure. Evaluation should therefore expose intermediate signals that connect prediction to action. Representative examples include state and contact changes, precondition satisfaction, uncertainty, intervention effects, failure causes, and recovery outcomes. The proposed \emph{Trace Card} organizes such information into a replayable physical record; its fields remain open to refinement through empirical use and community agreement.

\paragraph{Roadmap structure.}
The paper follows the argument above. Section~2 traces the overlapping evolution of action policies, semantic grounding, VLA models, and predictive world models toward WAMs. Section~3 explains why current model, data, and system advances remain difficult to accumulate. Section~4 first defines the roles and interfaces of the physical-intelligence stack. It then develops the embodied brain and WAM prediction contract, the physical harness and tool ecosystem, data and task records, closed-loop learning, ecosystem coordination, and near-term milestones. Section~5 summarizes how these interfaces support scalable reuse without prescribing a final architecture.

\paragraph{Contributions.}
This roadmap makes five contributions:
\begin{itemize}
\item \textbf{Scaling diagnosis.} We distinguish model capacity from capability reuse and system-level accumulation, then organize their limitations through three coupled gaps.
\item \textbf{WAM prediction contract.} We frame WAMs through decision-relevant consequence prediction rather than a mandatory output modality or final architecture.
\item \textbf{Embodied-brain system model.} We define the embodied brain as a long-term reasoning target and separate its intent interface from harness-mediated execution.
\item \textbf{Experience and evaluation contracts.} We introduce Embodiment, Task, and Trace Cards together with layered evaluation and update gates.
\item \textbf{Co-evolution roadmap.} We connect brain capabilities, representations, harnesses, tools, data, tasks, and post-training through testable interfaces and near-term milestones.
\end{itemize}

%% file: chapters/01a_visual_overview.tex
\begin{figure}[tp]
    \centering
    \includegraphics[width=0.75\linewidth]{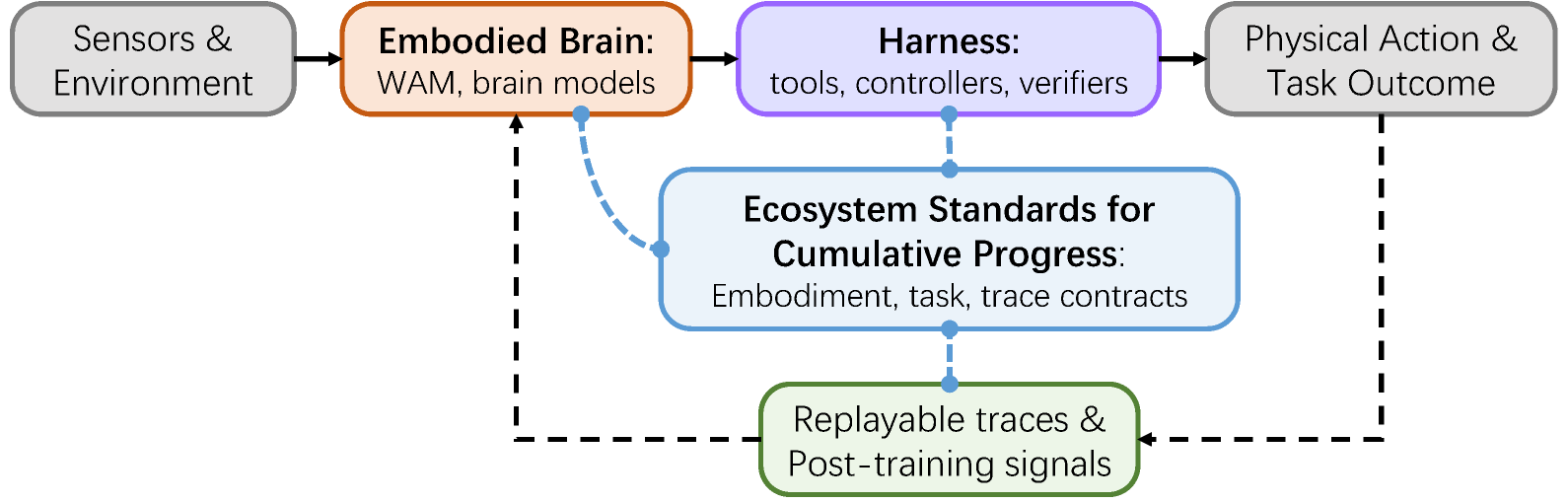}
\caption{System-level progression from WAM research to an embodied brain and a co-evolving physical-intelligence stack. WAMs provide a current route for studying intervention-conditioned prediction, while the embodied brain remains the broader target for reusable physical reasoning and intent formation.}
\label{fig:paper-thesis-overview}
\end{figure}
\begin{figure}[tp]
\centering
    \includegraphics[width=0.92\linewidth]{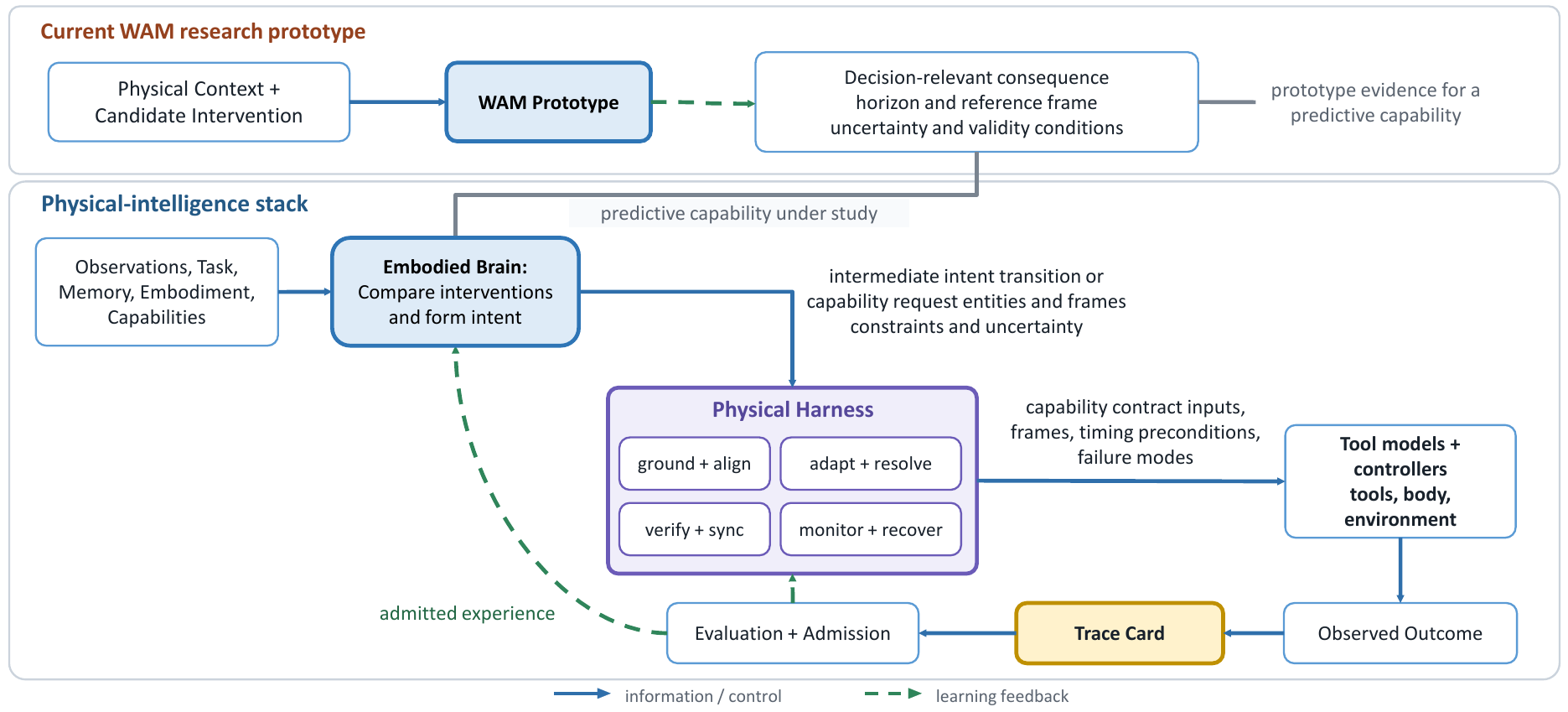}
\caption{Interfaces from prediction to physical execution. A WAM prototype estimates intervention consequences; the embodied brain forms intent; and the harness grounds capabilities, coordinates execution, and records outcomes in a Trace Card for evaluation and learning.}
\label{fig:system-interfaces}
\end{figure}
\begin{figure}[tp]
    \centering
    \includegraphics[width=0.55\linewidth]{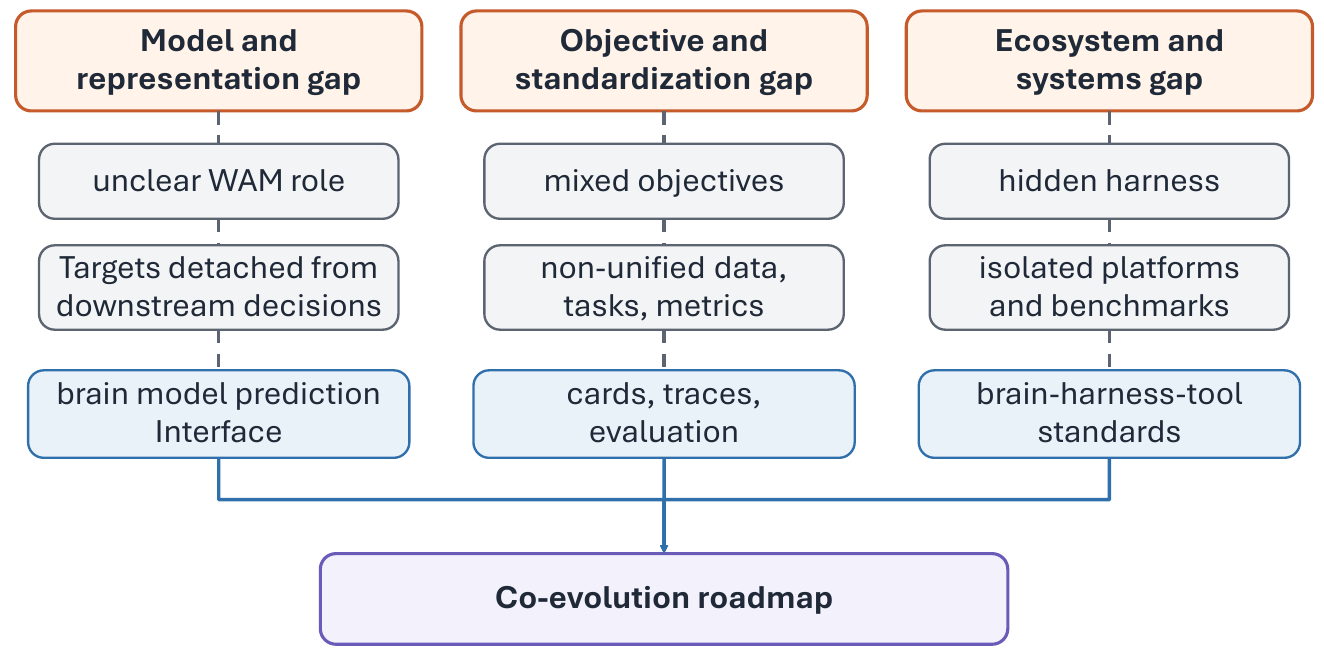}
\caption{Three coupled gaps and their roadmap responses. Model interfaces clarify prediction and intent; task, trace, and evaluation contracts support comparison; and shared brain--harness--tool standards coordinate system-level progress.}
\label{fig:three-gaps-to-roadmap}
\end{figure}

Figures~\ref{fig:paper-thesis-overview}--\ref{fig:three-gaps-to-roadmap} introduce the argument at three levels. Figure~\ref{fig:paper-thesis-overview} presents the system-level ambition, Figure~\ref{fig:system-interfaces} makes the model and execution interfaces explicit, and Figure~\ref{fig:three-gaps-to-roadmap} connects the three diagnosed gaps to the co-evolution roadmap.

%% file: chapters/02_evolution_toward_wams.tex
\section{Evolution of Embodied Models Toward WAMs}

The introduction frames WAMs as a step from passive perception and direct action prediction toward intervention-level physical foresight. This section makes that transition explicit. The evolution is not a single line in which action models become VLA models and then become WAMs. A more accurate account is a convergence of three trajectories. One learns executable robot actions, another grounds action in language and multimodal task context, and a third predicts how the world changes under intervention. WAMs become meaningful when these trajectories meet around a shared question: \emph{what will happen if this embodied system takes this action, invokes this skill, or calls this tool under the current physical constraints?}

\FloatBarrier

\subsection{Action Prediction as the First Interface}

One foundational trajectory in modern robot learning addresses a hard interface problem: transforming sensory and proprioceptive context into temporally coherent actions. Behavior Transformers model multimodal behaviors, ACT predicts action chunks for fine-grained bimanual manipulation, and Diffusion Policy represents visuomotor control as conditional action diffusion \citep{shafiullah2022behaviortransformerscloningk,liang2023maal,zhao2023learningfinegrainedbimanualmanipulation,chi2023diffusionpolicy}. QT-Opt demonstrates a complementary route in which large-scale real-robot interaction supports closed-loop visuomotor reinforcement learning \citep{kalashnikov2018qtoptscalabledeepreinforcement}. These works define executable action interfaces on which later embodied foundation models build.

The limitation is that the expected consequence of an action is often implicit. A policy may select a good motion without exposing which physical preconditions it relies on, which state transition it expects, how uncertainty affects the decision, or why a nearby alternative is unsafe. This is acceptable for narrow reactive skills, but it becomes a bottleneck for long-horizon manipulation, tool use, recovery, safety verification, and learning from failures. A system that only emits the next action is difficult to inspect, compare, and improve as an embodied brain.

\subsection{Sequence Modeling and Semantic Grounding}

A partly overlapping development made actions and trajectories compatible with foundation-model-style sequence learning. Decision Transformer casts reinforcement learning as conditional sequence modeling over returns, states, and actions, while Trajectory Transformer models trajectories for offline control \citep{chen2021decisiontransformerreinforcementlearning,janner2021offlinereinforcementlearningbig}. Gato extends a shared sequence-modeling interface across text, games, and embodied control \citep{reed2022generalistagent}. These systems do not define WAMs, but they establish that actions, states, rewards, language, and observations can be represented within related sequence interfaces.

In parallel, language-conditioned and multimodal systems expanded task specification. CLIPort combines semantic representations with spatial action localization, and BC-Z studies zero-shot task generalization from task-conditioned imitation \citep{shridhar2021cliportpathwaysroboticmanipulation,jang2022bczzeroshottaskgeneralization}. SayCan grounds language-model skill proposals in robotic affordances, while VIMA uses multimodal prompts for tabletop manipulation \citep{ahn2022saycan,jiang2022vima}. PaLM-E integrates continuous sensor inputs into an embodied multimodal language model \citep{driess2023palme}. Language can therefore specify goals, objects, and constraints. Physical success still depends on reachability, contact, geometry, force, and task progress under action.

\subsection{VLA Policies Scale Observation-to-Action Learning}

Building on these developments, VLA models connect large-scale vision-language priors with executable robot action spaces. RT-1 scales transformer policies on real-robot data, while RT-2 co-trains vision-language models with robot trajectories and expresses robot actions as token-compatible outputs \citep{brohan2022rt1,brohan2023rt2}. Open X-Embodiment and RT-X foreground cross-robot data aggregation; Octo and OpenVLA provide open generalist policy foundations; and $\pi_0$ uses flow matching for continuous action generation \citep{oneill2023openx,octo2024,kim2024openvla,black2024pi0}. Together, these systems establish a broad policy trajectory that maps multimodal context and instructions to actions across varied tasks and embodiments.

However, the VLA interface remains primarily action-oriented. A VLA may contain implicit physical knowledge, but the prediction is not necessarily exposed as a reusable contract. Downstream systems may not know which future states were considered, which constraints were violated, whether a recovery plan exists, or how confident the model is. This does not make VLA obsolete. Under the distinction adopted here, a VLA interface emphasizes executable action generation, whereas a WAM interface exposes consequence estimates that help compare, verify, revise, or reject candidate interventions.

\begin{figure*}[tp]
    \centering
    \includegraphics[width=\linewidth]{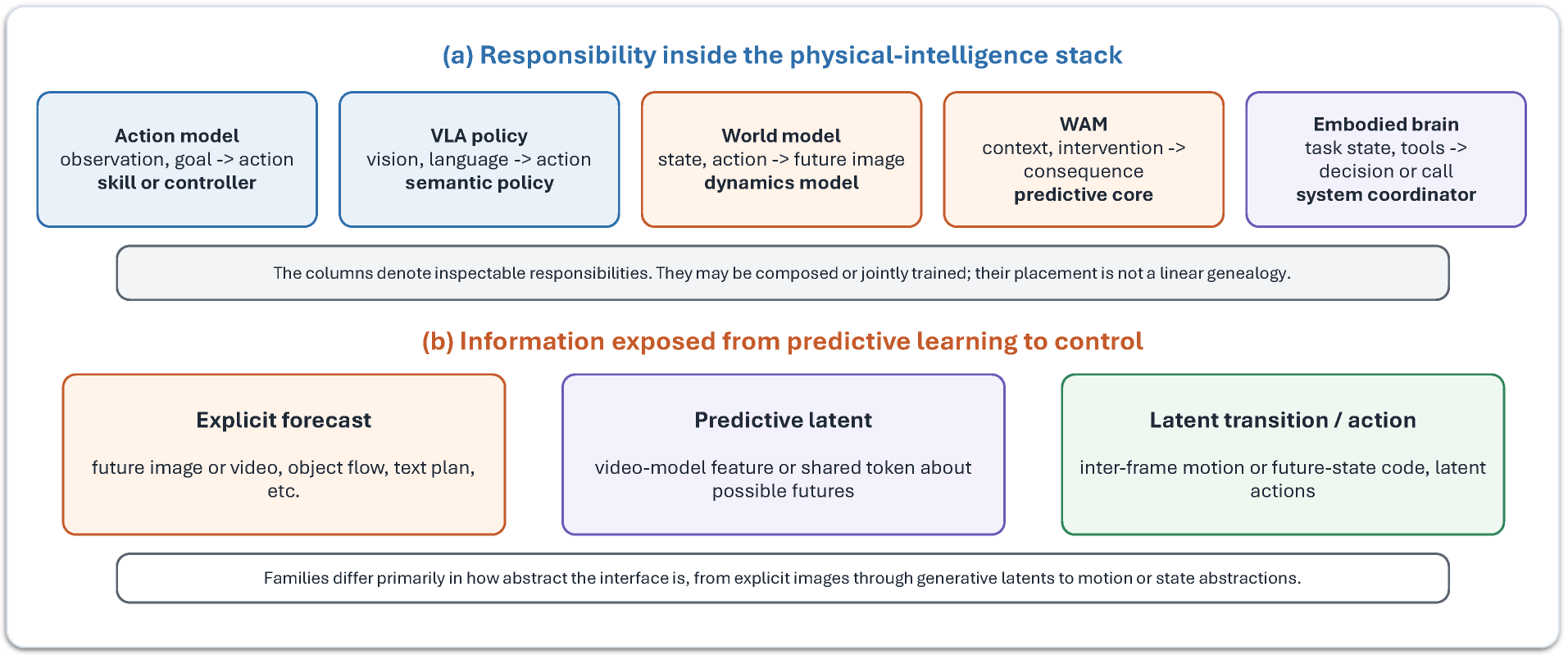}
\caption{Roles and predictive interfaces across embodied-model families. The upper panel distinguishes action models, VLA policies, world models, WAMs, and the embodied-brain target by their primary system responsibility. The lower panel compares explicit forecasts, predictive latent representations, and latent transition or action interfaces. These categories may overlap and do not imply a linear historical sequence.}
\label{fig:model-family-taxonomy}
\end{figure*}

Figure~\ref{fig:model-family-taxonomy} separates two questions that are often conflated: the responsibility of a model inside the larger system, and the representation through which predictive information reaches control. It is a conceptual taxonomy used by this paper, not a claim that current systems always obey clean boundaries.

\subsection{World Models Make Prediction Useful for Control}

A second trajectory begins from prediction. Action-conditioned video models learn future observations under candidate actions, while visual-foresight systems connect those predictions to model-predictive control \citep{finn2016unsupervisedlearningphysicalinteraction,finn2017deepvisualforesightplanning,ebert2018visualforesightmodelbaseddeep}. DreamerV3 learns a compact world model that predicts outcomes of potential actions; its actor and critic improve behavior using imagined latent trajectories rather than planning directly in raw pixel space \citep{hafner2023dreamerv3}. UniPi and RoboDreamer explore complementary uses of generated visual futures as planning or policy signals \citep{du2023learninguniversalpoliciestextguided,zhou2024robodreamerlearningcompositionalworld}.

The key lesson is not that robots must render future videos. The lesson is that prediction becomes useful when it is connected to intervention. A visually plausible rollout that does not distinguish feasible from infeasible actions is weak for decision making. Conversely, a compact latent state, contact estimate, affordance map, risk score, or task-progress prediction can be valuable if it changes what the system chooses to do next. This observation prepares the WAM view: output modality matters less than whether the predicted consequence constrains action.

\subsection{WAM-Oriented Models Couple Prediction and Control}

Recent systems connect prediction and control through increasingly diverse interfaces. GR-1 predicts future images and actions after video-generative pretraining, while GR-2 combines video, language, and action modeling \citep{wu2023gr1,cheang2024gr2generativevideolanguageactionmodel}. Prediction with Action and Unified Video Action Model jointly model visual futures and robot actions, whereas WLA integrates world modeling, language reasoning, and action synthesis \citep{guo2024predictionactionvisualpolicy,li2025unifiedvideoactionmodel,yang2026wla}. GWM provides a 3D Gaussian world-model example, showing that the predicted representation need not be restricted to RGB video \citep{lu2025gwm}. The WAM survey groups this broader literature into cascaded and joint designs, with further variation in generation modality, conditioning mechanism, and action decoding \citep{wang2026wam}.

These papers use different self-descriptions, including generative policy, VLA, unified video-action model, world-language-action model, and world model. We therefore refer to them collectively as \emph{WAM-oriented and closely related predictive-control models}; we do not retroactively relabel every method as a WAM. Their common relevance is narrower: each investigates how information about possible futures can influence action generation, planning, or control.

Taken together, the differing interfaces suggest that the field is still searching for an effective predictive contract. The position adopted in this paper is therefore deliberately provisional: WAMs are not universal video generators, replacements for VLAs, or complete embodied brains. They are current research prototypes for learning and exposing decision-relevant consequences under candidate interventions. Whether these capabilities will remain a distinct module, be absorbed into a more general brain model, or be reorganized in another architecture is an open question. The Roadmap later develops the capabilities and interfaces that this line of research should clarify; this section first examines what contemporary systems actually pass from prediction to control.

\begin{figure*}[tp]
    \centering
    \includegraphics[width=\linewidth]{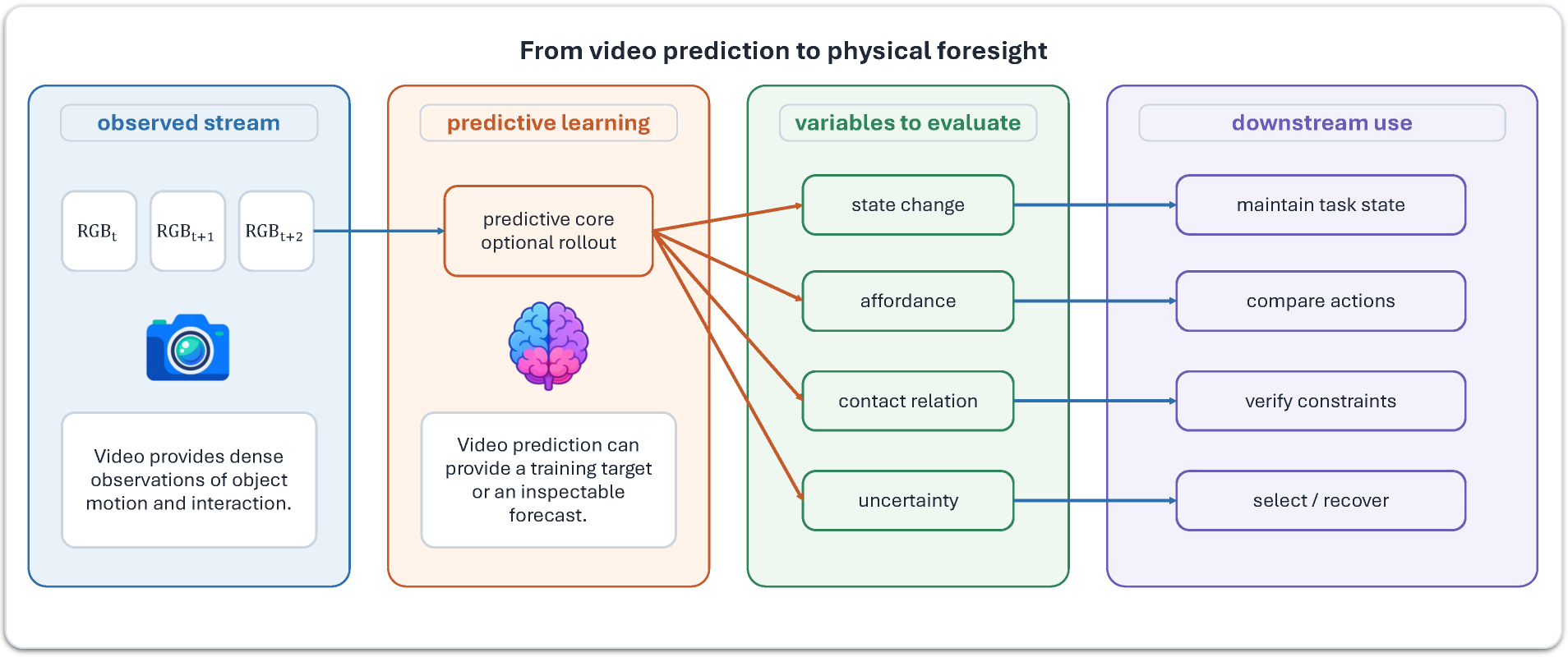}
\caption{Video prediction as one route to decision-relevant physical foresight. Observed frames and candidate interventions condition predictive learning; state change, affordance, contact, and uncertainty are evaluation targets rather than guaranteed outputs. Downstream modules may use validated predictions to imagine future states, compare actions, verify constraints, or select recovery behavior.}
\label{fig:from-video-to-physical-foresight}
\end{figure*}

Figure~\ref{fig:from-video-to-physical-foresight} summarizes the paper's interpretation of video prediction as one training and inspection route toward physical foresight rather than as the defining output of every WAM. The intermediate variables in the figure are candidate evaluation targets, not capabilities that follow automatically from video prediction.

\subsection{Contemporary Predictive Interfaces for Robot Control}

The historical trajectories above explain why prediction and action are now being studied together. A complementary question is how contemporary systems connect them. One useful analytical axis is the information made available at the predictive--control boundary. We distinguish three non-exclusive families: explicit observation or geometric forecasts, predictive latent representations, and learned transition or action abstractions. This taxonomy describes an interface, not the name chosen by each paper and not a performance ordering.

\paragraph{Explicit observation or geometric foresight.}
The most inspectable interfaces materialize an intermediate plan or forecast before execution. UniPi generates a task-conditioned visual trajectory and learns an inverse-dynamics mapping from predicted visual goal to actions, while VLP searches over video and language plans and executes intermediate visual goals through goal-conditioned policies \citep{du2023learninguniversalpoliciestextguided,du2024videolanguageplanning}. AVDC synthesizes a task execution and combines inter-frame correspondences with depth and geometric control, avoiding action annotations for the demonstration videos \citep{ko2024actionlessvideos}. Gen2Act instead generates human task videos and trains a robot policy to follow the resulting visual motion specification \citep{bharadhwaj2025gen2act}.

Explicit interfaces need not be RGB rollouts. Im2Flow2Act predicts object flow from human video and uses that flow to condition a policy trained with simulated robot play \citep{xu2025im2flow2act}. The $\pi_{0.7}$ work presents a steerable generalist robotic foundation model rather than an explicit-foresight architecture. Its multimodal prompt may nevertheless include optional visual subgoals, and the associated project report notes that such subgoals can be synthesized by a lightweight world model \citep{physicalintelligence2026pi07paper,physicalintelligence2026pi07}. We include it here only as an example of a predictive artifact exposed to a control policy. These examples illustrate the value of inspectable intermediate targets, while also showing that converting a forecast into robot commands remains a separate design problem.

\paragraph{Predictive latent representations.}
A second family uses internal features of predictive video or joint world-action models without requiring a fully decoded visual rollout at every control step. VPP conditions an inverse-dynamics policy on predictive representations inside a video diffusion model \citep{hu2025videopredictionpolicy}. The mimic-video similarly connects a pretrained video model to a flow-matching action decoder through latent representations \citep{pai2025mimicvideo}. These designs may avoid some decoding cost, but their physical content is less directly inspectable and must be established through downstream ablations and control experiments rather than assumed from the representation alone.

Joint models provide another form of latent exchange. UWM combines video and action diffusion in a single transformer and uses modality-specific diffusion timesteps to instantiate policy, forward-dynamics, inverse-dynamics, or video-generation behavior \citep{zhu2025unifiedworldmodels}. RynnVLA-002 jointly predicts future images and robot actions in a unified VLA/world-model framework \citep{cen2025rynnvla002}. They show how prediction and action can share parameters or tokens, but they do not establish that a single unified architecture is preferable across embodiments or control regimes.

\paragraph{Latent transition or action abstractions.}
A third family learns a compact code for change between observations. LAPA first quantizes inter-frame changes into discrete latent actions, then pretrains a VLA to predict these codes and fine-tunes the model with action-labeled robot data to map them to physical commands \citep{ye2025latentactionpretraining}. The villa-X extends latent-action modeling within VLA pretraining, while VLA-JEPA predicts future latent states from the current observation before fine-tuning an action head \citep{chen2025villax,sun2026vlajepa}. A latent action is therefore not automatically a robot action: it may encode motion, viewpoint change, or other visual variation, and its correspondence to an embodiment-specific control space must be learned and evaluated.

Across the three families, dynamics learning and action grounding are separated to different degrees. Some methods first learn predictive structure from action-free or weakly labeled video and later connect that representation to a robot through action-labeled data, geometry, or an existing controller. Others optimize prediction and action jointly. The interface can therefore range from explicit pixels, flow, or text to predictive latents and purpose-built transition codes. This interface choice is orthogonal to whether the complete architecture is modular or jointly trained.

Table~\ref{tab:wam-interface-landscape} compares representative systems along this interface axis. It records each method's own framing, its connection to control, the role of its data, and the scope of its reported evaluation.

\begin{table}[tp]
\centering
\scriptsize
\setlength{\tabcolsep}{2.0pt}
\renewcommand{\arraystretch}{1.10}
\begin{tabularx}{\textwidth}{P{0.14\textwidth} P{0.18\textwidth} P{0.20\textwidth} P{0.22\textwidth} Y}
\toprule
\textbf{System} & \textbf{Exposed interface} & \textbf{Control coupling} & \textbf{Reported data role} & \textbf{Evaluation scope} \\
\midrule
\rowcolor{figgray!8}\multicolumn{5}{l}{\textit{Explicit observation or geometric foresight}} \\
UniPi \citep{du2023learninguniversalpoliciestextguided} & Pixel-space visual trajectory & Inverse dynamics maps visual transitions to actions & Generative pretraining plus action mapping & Simulated and real manipulation \\
VLP \citep{du2024videolanguageplanning} & Generated video and text plan & Tree search plus goal-conditioned policies & Pretrained video/VLM components and task data & Long-horizon simulated and real tasks \\
Actionless Videos \citep{ko2024actionlessvideos} & Execution video and dense correspondence & Geometry, inverse kinematics, and motion planning & RGB demonstrations without robot commands & Simulated and real manipulation and navigation \\
Gen2Act \citep{bharadhwaj2025gen2act} & Generated human task video & Closed-loop video-conditioned robot policy & Video generator plus robot interaction data & Real-robot manipulation \\
Im2Flow2Act \citep{xu2025im2flow2act} & Predicted object flow & Flow-conditioned policy & Human video plus simulated robot play & Real rigid, articulated, and deformable-object tasks \\
F1 \citep{lv2025f1} & Goal-conditioned visual foresight & Foresight-guided inverse dynamics & Multi-stage robot-trajectory training & Simulation and real-robot studies \\
$\pi_{0.7}$\textsuperscript{\dag} \citep{physicalintelligence2026pi07paper} & Multimodal prompt; optional visual subgoals & High-level policy and action expert & Robot, human, and autonomous interaction data & Multi-platform real-robot studies \\
\midrule
\rowcolor{figgray!8}\multicolumn{5}{l}{\textit{Predictive latent representations and joint video--action models}} \\
VPP \citep{hu2025videopredictionpolicy} & Predictive video-model features & Inverse dynamics conditioned on predictive features & Human and robot manipulation video & CALVIN, simulation, and real dexterous manipulation \\
mimic-video \citep{pai2025mimicvideo} & Latent video-space plan & Flow-matching action decoder & Video pretraining plus robot trajectories & Simulated and real manipulation \\
UWM \citep{zhu2025unifiedworldmodels} & Joint video/action diffusion & Policy, forward/inverse dynamics, or joint generation & Labeled trajectories and robot video with labels withheld & LIBERO and real Franka tasks \\
RynnVLA-002 \citep{cen2025rynnvla002} & Shared action and future-image prediction & Joint action generation and prediction & Robot task data & LIBERO and real-robot tasks \\
\midrule
\rowcolor{figgray!8}\multicolumn{5}{l}{\textit{Latent transition or action abstractions}} \\
LAPA \citep{ye2025latentactionpretraining} & Discrete inter-frame latent action & Action-labeled tuning maps codes to commands & Video without used robot commands, then robot grounding & SIMPLER and real-robot manipulation \\
villa-X \citep{chen2025villax} & Learned latent action & Latent-action planning with robot-action learning & Robot and visual-transition data & SIMPLER, LIBERO, and real-robot settings \\
VLA-JEPA \citep{sun2026vlajepa} & Predicted future latent state & Action-head tuning after latent pretraining & Video pretraining followed by robot-action data & LIBERO-family, SIMPLER, and real-robot tasks \\
\bottomrule
\end{tabularx}
\vspace{2pt}

\parbox{\textwidth}{\footnotesize \textsuperscript{\dag}$\pi_{0.7}$ is a VLA. It appears in the explicit-interface group only because its multimodal prompt may include visual subgoals; this placement does not characterize the overall architecture as an explicit-foresight model.}
\caption{Representative predictive-control interfaces. Evaluation settings are not directly comparable across different robots, tasks, and protocols.}
\label{tab:wam-interface-landscape}
\end{table}

\subsection{Data and Supervision Landscape}


The model taxonomy is inseparable from the data used to learn each interface. On one hand, action-labeled real-robot trajectory demand physical execution and often teleoperation. On the other hand, action-free sources including human and internet videos can provide broader observations of interaction without commands in the target robot's action space \citep{khazatsky2024droid,ye2025latentactionpretraining,bharadhwaj2025gen2act}. This asymmetry motivates several predictive-control methods, yet its practical relevance is largely contingent on how each method handles the missing action information.

\begin{figure*}[tp]
    \centering
    \includegraphics[width=\linewidth]{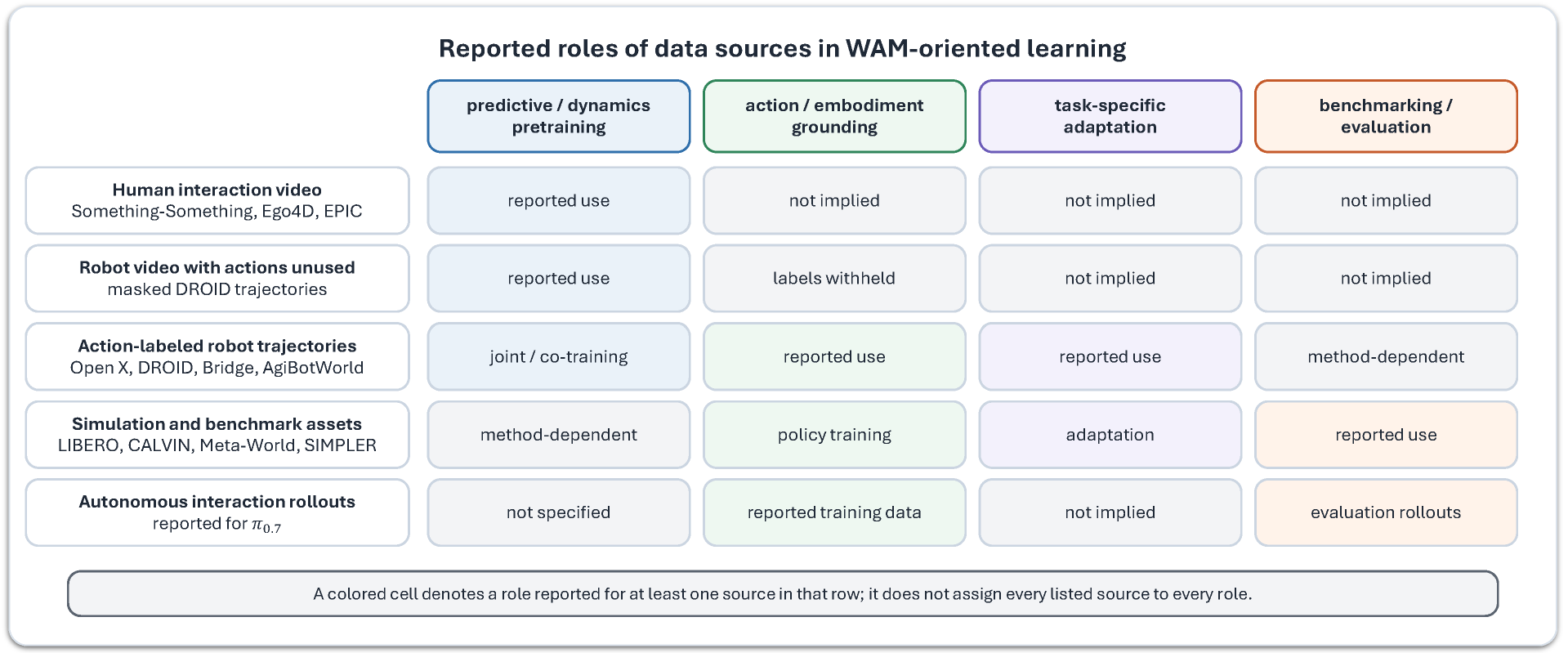}
\caption{Reported roles of data sources in WAM-oriented learning. Cells summarize representative or method-dependent uses documented in the cited literature; they do not imply that every resource in a row supports every learning role. The simulation and benchmark row is a coarse grouping whose individual resources may serve training, adaptation, or evaluation differently.}
\label{fig:wam-data-supervision}
\end{figure*}

Different types of data can be employed at various training stages. Human video may support predictive or latent pretraining. Robot trajectories may support prediction and action grounding jointly, while simulation may provide training, adaptation, or evaluation. UWM studies both action-labeled DROID trajectories and robot videos for which action annotations are withheld. LAPA instead learns a visual transition code before grounding it with robot actions \citep{zhu2025unifiedworldmodels,ye2025latentactionpretraining}. Figure~\ref{fig:wam-data-supervision} therefore organizes sources by reported learning role rather than imposing one pretrain--fine-tune sequence.

Autonomous interaction introduces another role. The $\pi_{0.7}$ paper and project report describe policy-generated rollouts, including suboptimal evaluation data, as training data conditioned on quality, speed, and related metadata \citep{physicalintelligence2026pi07paper,physicalintelligence2026pi07}. Such data can expose failures and recovery opportunities that expert demonstrations omit. Its usefulness nevertheless depends on trace quality, outcome attribution, and safeguards against reinforcing model errors.

\paragraph{Human video.}
Collections vary substantially in viewpoint and annotation. \citet{goyal2017something} introduced the original benchmark; the official V2 release contains 220,847 short human--object interaction clips \citep{qualcomm2026ssv2}. LAPA and VPP use this dataset in their respective video-learning pipelines \citep{ye2025latentactionpretraining,hu2025videopredictionpolicy}. Ego4D and EPIC-KITCHENS-100 provide longer egocentric recordings with different activity and annotation structures \citep{grauman2021ego4d,damen2022epickitchens100}. These sources can broaden motion and semantic coverage, but they do not expose target-robot commands. They also contain camera motion, partial observability, and human-specific kinematics that a robot learner must account for.

Here, \emph{action-free} does not necessarily mean \emph{unannotated}. Human-video datasets may provide action categories, narrations, temporal segments, or other semantic labels. What they lack is a time-aligned command stream in the target robot's action space. Semantic supervision can therefore support representation learning without resolving how a depicted transition should be executed by a particular embodiment.

\paragraph{Robot trajectories.}
Robot datasets provide more direct action grounding, but they remain heterogeneous. Open X-Embodiment aggregates data from 22 robot embodiments, while the final RSS paper for DROID reports 76,000 trajectories across 564 scenes and 86 tasks \citep{oneill2023openx,khazatsky2024droid}. BridgeData V2 contains 60,096 trajectories across 24 environments \citep{bridgedatav2}. AgiBotWorld Colosseo subsequently reports more than one million trajectories across 217 tasks and five scenarios \citep{bu2025agibotworld}. Scale alone does not resolve differences in sensing, action spaces, control frequencies, task definitions, or collection policies.

Robot supervision is also richer than an action vector. Depending on the resource, a trajectory may include language instructions, proprioceptive state, multiple camera streams, depth, calibration, or related metadata \citep{oneill2023openx,khazatsky2024droid,bridgedatav2}. These fields determine which predictive targets can be learned and whether results can be compared across datasets. Additional modalities can improve observability, but compatibility still requires explicit conventions and mappings.

Public datasets should also be distinguished from paper-specific collections. VPP reports self-collected Panda-arm and dexterous-hand data, while Gen2Act combines a pretrained video generator with robot interaction data collected for its policy \citep{hu2025videopredictionpolicy,bharadhwaj2025gen2act}. Such collections provide embodiment-specific grounding, but their scale, schema, and availability must be read from the corresponding source. They do not support a field-wide claim that real-robot data are uniformly public or proprietary.

\paragraph{Simulation and evaluation resources.}
Simulation assets play several roles. LIBERO provides four lifelong-learning suites with 130 manipulation tasks and teleoperated demonstrations, while CALVIN targets long-horizon language-conditioned manipulation \citep{liu2023libero,mees2022calvin}. Meta-World provides 50 simulated manipulation tasks \citep{yu2020metaworld}. SIMPLER should be distinguished from these training corpora: it provides simulated environments and a workflow for evaluating real-world manipulation policies through paired simulation and real-robot studies \citep{li2024simpler}. Table~\ref{tab:wam-data-landscape} records these differences without imposing a binary public/proprietary split.

\begin{table}[tp]
\centering
\scriptsize
\setlength{\tabcolsep}{2.2pt}
\renewcommand{\arraystretch}{1.02}
\begin{tabularx}{\textwidth}{P{0.17\textwidth} P{0.19\textwidth} P{0.19\textwidth} P{0.25\textwidth} Y}
\toprule
\textbf{Resource} & \textbf{Source / supervision} & \textbf{Verified scale or scope} & \textbf{Documented use or relevance} & \textbf{Artifact availability} \\
\midrule
\rowcolor{figgray!8}\multicolumn{5}{l}{\textit{Human interaction and egocentric video}} \\
Something-Something V2 \citep{qualcomm2026ssv2} & Short human-object videos; labels but no robot commands & 220,847 clips in the official V2 release & Predictive pretraining in VPP and latent-transition pretraining in LAPA \citep{hu2025videopredictionpolicy,ye2025latentactionpretraining} & Research dataset; access terms apply \\
Ego4D \citep{grauman2021ego4d} & Egocentric human video with multiple annotations & 3,670 hours across 74 locations & Broad egocentric visual experience; not a robot-action corpus & Public research dataset with access agreement \\
EPIC-KITCHENS-100 \citep{damen2022epickitchens100} & Unscripted egocentric kitchen video & 100 hours, 90k action segments, 45 kitchens & Fine-grained human-object interaction and anticipation research & Public research dataset with access terms \\
\midrule
\rowcolor{figgray!8}\multicolumn{5}{l}{\textit{Robot-centric trajectories}} \\
Open X-Embodiment \citep{oneill2023openx} & Aggregated multi-robot trajectories with actions & 22 robot embodiments and 527 skills & Cross-embodiment policy learning and model pretraining & Released in standardized dataset formats \\
DROID \citep{khazatsky2024droid} & Real Franka demonstrations with synchronized observations and actions & 76k trajectories, 350 hours, 564 scenes, 86 tasks & Action grounding and action-label withholding in UWM \citep{zhu2025unifiedworldmodels} & Dataset, code, and hardware guide released \\
BridgeData V2 \citep{bridgedatav2} & Real WidowX manipulation trajectories & 50,365 demonstrations plus 9,731 scripted rollouts across 24 environments & Multi-task policy learning and latent-transition pretraining in LAPA \citep{ye2025latentactionpretraining} & Dataset and pretrained models released \\
AgiBotWorld Colosseo \citep{bu2025agibotworld} & Large-scale multi-scenario robot trajectories & More than 1M trajectories, 217 tasks, five scenarios & Generalist policy and latent-action research & Dataset, tools, and models announced as open artifacts \\
Autonomous interaction data \citep{physicalintelligence2026pi07paper} & Policy-generated robot rollouts with multimodal context & Scale not publicly specified & One reported component of $\pi_{0.7}$ generalist training & No standalone public dataset claimed here \\
\midrule
\rowcolor{figgray!8}\multicolumn{5}{l}{\textit{Simulation training and benchmark assets}} \\
LIBERO \citep{liu2023libero} & Simulated manipulation plus teleoperated demonstrations & Four suites, 130 tasks & Lifelong transfer, policy training, and evaluation & Benchmark, code, and demonstrations released \\
CALVIN \citep{mees2022calvin} & Language-conditioned simulated manipulation & Long-horizon sequences in multiple environments & Policy learning and compositional task evaluation & Environment, benchmark, and baselines released \\
Meta-World \citep{yu2020metaworld} & Simulated multi-task manipulation & 50 distinct tasks & Multi-task and meta-RL training/evaluation & Open-source benchmark \\
\midrule
\rowcolor{figgray!8}\multicolumn{5}{l}{\textit{Evaluation frameworks}} \\
SIMPLER \citep{li2024simpler} & Simulated environments paired with real-policy evaluation & Common real-robot setups and distribution shifts & Evaluation framework, not a generic fine-tuning dataset & Environments and creation workflow released \\
\bottomrule
\end{tabularx}
\captionsetup{skip=4pt}
\caption{Data resources used in or relevant to predictive-control and WAM-oriented research. Scale, scope, and availability follow the primary sources and do not imply a common license, schema, action space, or annotation level.}
\label{tab:wam-data-landscape}
\end{table}

The resulting data picture is more nuanced than a substitution of video for robot trajectories. Broad video may reduce dependence on action labels for selected objectives, but converting a learned representation into robot commands still requires an embodiment-specific mapping. Depending on the method, this mapping may be learned from robot data or supplied through simulation, geometry, inverse kinematics, or an existing controller. Conversely, action-labeled scale does not create a reusable predictive interface when embodiment metadata, sensing conventions, and evaluation protocols remain incompatible. The central problem is therefore not data diversity itself, but the absence of interfaces that allow diverse experience to accumulate.

\subsection{Evaluation Across Prediction, Grounding, and Control}

Evaluation also varies with the interface being studied. The WAM survey organizes emerging protocols around visual fidelity, physical commonsense, and action plausibility \citep{wang2026wam}. Individual model papers add representation ablations, inverse-dynamics or action-generation tests, and closed-loop task success in different simulated and real settings \citep{du2024videolanguageplanning,hu2025videopredictionpolicy,ye2025latentactionpretraining,zhu2025unifiedworldmodels}. These measurements answer different questions and should not be collapsed into one leaderboard.

RoboWM-Bench makes the distinction especially clear. It converts behavior depicted in generated manipulation videos into embodied action sequences and validates them through robot execution, reporting failures involving spatial reasoning, contact prediction, and non-physical deformation \citep{robowmbench2026}. This protocol tests whether visual behavior can be made executable; it does not imply that visual fidelity, representation quality, action accuracy, and end-task success are interchangeable. Table~\ref{tab:wam-evaluation-landscape} summarizes four complementary evidence layers found in the literature.

\begin{table}[tp]
\centering
\scriptsize
\setlength{\tabcolsep}{3pt}
\renewcommand{\arraystretch}{1.08}
\begin{tabularx}{\textwidth}{>{\columncolor{figgray!5}}P{0.18\textwidth} P{0.23\textwidth} P{0.31\textwidth} Y}
\toprule
\textbf{Evidence layer} & \textbf{Question} & \textbf{Reported evidence and representative sources} & \textbf{What it does not establish} \\
\midrule
Predictive fidelity and physical consistency & Does the predicted future resemble the target and preserve plausible spatial or physical change? & Video metrics and qualitative rollouts in predictive models such as VPP; visual-fidelity and physical-commonsense categories in the WAM survey \citep{hu2025videopredictionpolicy,wang2026wam} & That the forecast can be converted into an executable or useful robot action \\
Interface and action grounding & Does the intermediate forecast or latent improve action inference under the tested embodiment? & Inverse-dynamics, correspondence, flow-conditioning, and latent-action studies in UniPi, Actionless Videos, Im2Flow2Act, LAPA, and UWM \citep{du2023learninguniversalpoliciestextguided,ko2024actionlessvideos,xu2025im2flow2act,ye2025latentactionpretraining,zhu2025unifiedworldmodels} & A common action space, transfer guarantee, or calibrated consequence estimate \\
Closed-loop task utility & Does the complete policy solve the task under its simulation or real-robot protocol? & Task success, generalization splits, and long-horizon evaluations reported by VLP, Gen2Act, VPP, and related systems \citep{du2024videolanguageplanning,bharadhwaj2025gen2act,hu2025videopredictionpolicy} & Which component caused success or failure, or whether results compare across platforms \\
Embodiment-grounded executability & Can behavior inferred from a generated future be translated into actions that work on a robot? & Generated-behavior conversion and execution in RoboWM-Bench \citep{robowmbench2026} & Full embodied-brain reasoning, harness quality, uncertainty calibration, or continual improvement \\
\bottomrule
\end{tabularx}
\captionsetup{skip=4pt}
\caption{Complementary evidence for predictive-control evaluation. The diagnostic dimensions are not a universal hierarchy; uncertainty, traceability, harness behavior, safety, and update admission remain prospective extensions.}
\label{tab:wam-evaluation-landscape}
\end{table}

Three unresolved dependencies emerge from this review. First, contemporary methods expose different predictive variables, leaving both the physical representation and the eventual brain-model output interface unsettled. Second, data sources, supervision schedules, and evaluation protocols remain difficult to compare, creating an objective and standardization problem. Third, the mechanism that consumes a prediction---an inverse model, policy, planner, controller, or larger runtime---is often method-specific. A representation learned by one model may therefore be unusable by another component without a dedicated adapter or additional training. These observations connect the evolutionary landscape directly to the three gaps analyzed in Section~3.

Table~\ref{tab:evolution-genealogy} complements the timeline by comparing the primary interface, capability, and remaining limitation of each model family. The entries summarize the cited literature and the paper's interface-oriented interpretation; they are not intended as mutually exclusive categories.

\begin{table}[tp]
\centering
\scriptsize
\setlength{\tabcolsep}{2pt}
\renewcommand{\arraystretch}{1.08}
\begin{tabularx}{\textwidth}{>{\columncolor{figgray!5}}P{0.16\textwidth} P{0.20\textwidth} P{0.20\textwidth} P{0.18\textwidth} Y}
\toprule
\textbf{Interface stage} & \textbf{Representative systems} & \textbf{Primary interface} & \textbf{What becomes possible} & \textbf{Remaining limitation} \\
\midrule
Action policies & QT-Opt, Behavior Transformer, ACT, Diffusion Policy & Observation and goal $\rightarrow$ action sequence & Smooth manipulation, multimodal behavior, scalable imitation and RL & Consequences, uncertainty, preconditions, and failure causes are usually implicit. \\
Semantic grounding & CLIPort, BC-Z, SayCan, VIMA, PaLM-E & Language or prompt $\rightarrow$ skill, plan, or action & Task semantics, object grounding, affordance-aware skill selection & Grounded language does not by itself guarantee physical foresight or contact reasoning. \\
VLA policies & RT-1, RT-2, RT-X, Octo, OpenVLA, $\pi_0$ & Vision-language context $\rightarrow$ action tokens or continuous actions & Large-scale multimodal and cross-embodiment priors linked to executable robot action & Intermediate prediction and system-level interfaces may remain implicit. \\
World models & Visual foresight, DreamerV3, UniPi, RoboDreamer & Candidate action or task $\rightarrow$ future state, latent, or video & Planning, imagination, model-predictive control, value learning & Prediction may be hard to consume as a standardized embodied decision contract. \\
WAM-oriented predictive-control models & GR-1, GR-2, Prediction with Action, UVA, WLA, GWM, UWM & Context or action $\rightarrow$ future observation, action, predictive latent, geometric forecast, or 3D state & Prediction and action can be trained, queried, or evaluated within a connected control pipeline & No shared contract yet specifies which consequences should be exposed across embodiments and downstream systems. \\
\bottomrule
\end{tabularx}
\caption{Detailed reference taxonomy of embodied-model families and control interfaces. Categories may overlap and do not imply a mandatory historical sequence; reported outputs remain narrower than the proposed consequence contract.}
\label{tab:evolution-genealogy}
\end{table}

%% file: chapters/03_current_barriers.tex
\section{Current Barriers on the Path to an Embodied Brain}

The previous section traced overlapping trajectories from action policies, semantic grounding, and world modeling toward WAM-oriented predictive control. These developments provide useful capabilities, yet they also expose three forms of entanglement. Model-level reasoning is often tied to a particular prediction or action interface. Learning signals are tied to project-specific data and task protocols. Execution is tied to a custom runtime and embodiment. We organize these limitations through the three gaps introduced in the Introduction: a \emph{model and representation gap}, an \emph{objective and standardization gap}, and an \emph{ecosystem and systems gap}. The purpose is to explain why progress remains difficult to compare, reuse, and accumulate.

\subsection{The Three Gaps Are Coupled}

The three gaps form one system problem. A representation cannot be evaluated without specifying which decision it should support. An objective cannot be interpreted without the data schema, action interface, and embodiment that produced its supervision. Execution infrastructure is not neutral when it determines how outputs are translated, which actions are feasible, and which failures return to learning. Model design, data construction, task definition, execution, and evaluation are therefore coupled through their interfaces.

This coupling explains why rapid progress does not always become cumulative capability. VLA policies map rich multimodal context to actions. WAM-oriented methods expose visual, latent, geometric, or joint world--action predictions. Datasets broaden task and embodiment coverage, while benchmarks test different aspects of reasoning, executability, and safety. These advances remain difficult to combine when their variables and system assumptions are implicit. A roadmap must therefore specify both what capability is being developed and how that capability can be reused.

\begin{figure*}[tp]
    \centering
    \includegraphics[width=0.7\linewidth]{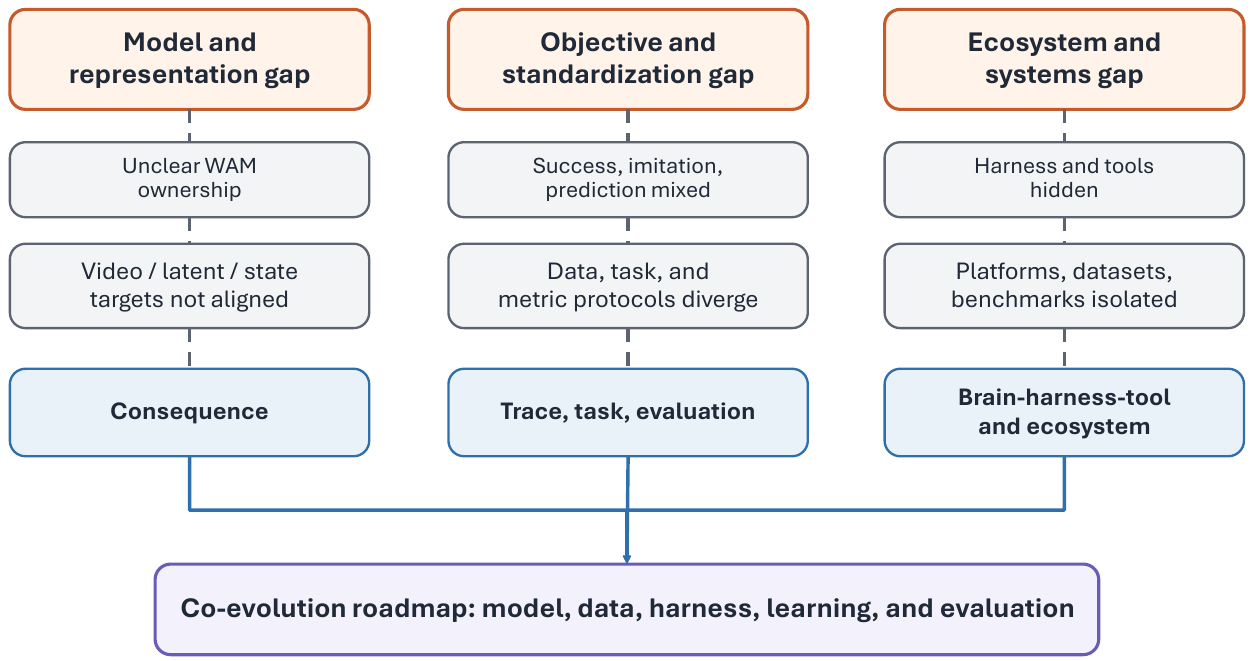}
\caption{Three coupled gaps that limit cumulative progress toward an embodied brain. Each column connects observable symptoms to a corresponding roadmap requirement: an explicit model interface, comparable task and experience contracts, or shared interfaces across the brain, harness, tools, and ecosystem.}
\label{fig:gap-roadmap-map}
\end{figure*}

Figure~\ref{fig:gap-roadmap-map} operationalizes the three introduction gaps as observable symptoms and corresponding contract requirements. It is the organizing synthesis for this chapter rather than an empirical taxonomy claimed by any single cited work.

\subsection{Model and Representation Gap}

The model and representation gap concerns what a general physical reasoning model should learn, expose, and leave to downstream execution. Current systems combine language and vision-language models, action tokenizers, diffusion policies, video generators, world models, and multimodal backbones. These components improve semantic grounding, action generation, or predictive modeling. They do not automatically yield a persistent and actionable representation of physical state. A model may imitate behavior or generate plausible video while still failing to track geometry, contact, uncertainty, or recovery conditions~\citep{liang2022seeg,liang2019vrr,liu2025uniinter,wang2026wam,robowmbench2026}.

Action-centric interfaces make the ownership problem concrete. Open X-Embodiment aligns data from many robots but retains embodiment-dependent normalization and control semantics \citep{oneill2023openx}. Octo supports new action spaces through adaptable output heads and fine-tuning \citep{octo2024}. These systems demonstrate valuable cross-embodiment transfer. They also show that an executable action remains meaningful only relative to a declared embodiment and controller. When the model output is inseparable from that local action interface, changing the hardware can require changes to the model head, training data, or both.

WAM research opens a complementary direction by exposing predicted consequences. GR-1 and GR-2 use video-generative learning for manipulation; Prediction with Action and UVA couple visual futures with action inference; and WLA combines world modeling, language reasoning, and action synthesis \citep{wu2023gr1,cheang2024gr2generativevideolanguageactionmodel,guo2024predictionactionvisualpolicy,li2025unifiedvideoactionmodel,yang2026wla}. Their outputs and training arrangements differ. The field has therefore not converged on a shared consequence interface that another model, planner, verifier, or controller can consume.

This gap is not equivalent to the ability to render video. Video offers dense temporal supervision and an interpretable forecast, but visual fidelity is not decision relevance. Useful physical reasoning may instead require pose, contact, traversability, preconditions, uncertainty, or expected progress. A WAM prototype can be valuable without photorealistic output when its representation improves decisions. Conversely, a plausible rollout provides limited evidence when it does not improve action selection, verification, recovery, or learning.

The embodied brain provides the model-level target, not a prescribed WAM module. It should integrate task context and physical evidence, compare possible interventions, and communicate an intended state transition or capability request. Current WAMs provide tractable prototypes for intervention-conditioned prediction without determining the final architecture. Embodiment-specific execution remains downstream: the harness grounds the brain output, while tool models and controllers realize the requested capability. This role boundary allows model capacity and execution capacity to improve through a declared interface.

\begin{figure*}[tp]
\centering
\resizebox{0.75\textwidth}{!}{
\begin{tikzpicture}[
    every node/.style={font=\sffamily\scriptsize},
    cause/.style={figwam, minimum width=2.55cm, minimum height=0.72cm},
    effect/.style={figcontract, minimum width=2.45cm, minimum height=0.66cm},
    fix/.style={figmodel, minimum width=2.70cm, minimum height=0.72cm}
]
\node[cause] (c0) at (0,0) {Unclear model\\ownership};
\node[effect] (c1) at (3.0,0.9) {mixed prediction\\targets};
\node[effect] (c2) at (3.0,0) {uncertain action\\interface};
\node[effect] (c3) at (3.0,-0.9) {hidden failure\\assumptions};
\node[effect] (c4) at (6.25,0.9) {incompatible\\objectives};
\node[effect] (c5) at (6.25,0) {weak evaluation\\attribution};
\node[effect] (c6) at (6.25,-0.9) {low trace\\reuse};
\node[fix] (fix) at (9.65,0) {declared interfaces:\\consequence, intent,\\capability, experience};
\draw[figdependency] (c0) -- (c1);
\draw[figdependency] (c0) -- (c2);
\draw[figdependency] (c0) -- (c3);
\draw[figdependency] (c1) -- (c4);
\draw[figdependency] (c2) -- (c5);
\draw[figdependency] (c3) -- (c6);
\draw[figarrow] (c4) -- (fix);
\draw[figarrow] (c5) -- node[candidateannotation, above] {motivates} (fix);
\draw[figarrow] (c6) -- (fix);
\end{tikzpicture}}
\caption{Analytical dependencies associated with unclear model ownership. Gray dashed lines are undirected and do not assert causality; blue arrows map the diagnosed limitations to declared consequence, intent, capability, and experience interfaces.}
\label{fig:barrier-dependency-map}
\end{figure*}
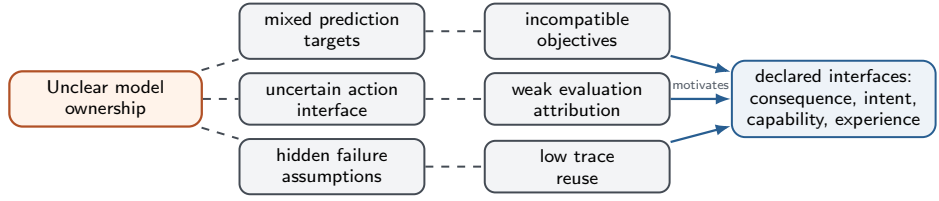

Figure~\ref{fig:barrier-dependency-map} summarizes how unclear ownership co-occurs with comparison and reuse problems. The connectors indicate dependencies in the paper's analysis, not empirically established causal effects.

\subsection{Objective and Standardization Gap}

The objective and standardization gap concerns what the field optimizes and whether learning signals remain meaningful across systems. Isolated task success is insufficient for this purpose. A system may succeed because the task distribution is narrow, the controller is strong, or recovery is unnecessary. It may fail because of incorrect state tracking, weak contact prediction, poor calibration, or an unavailable tool precondition. A final success rate does not distinguish these causes.

Training objectives also target different capabilities. Imitation learning can produce competent actions without exposing foresight. Future prediction can model plausible dynamics without selecting a better intervention. Reinforcement learning can improve reward while exploiting shortcuts, and verifier labels depend on the quality of their feedback source. These signals should not be collapsed into a single measure. Representation learning, consequence prediction, interface grounding, tool-chain supervision, verifier alignment, and closed-loop post-training require different evidence.

The deeper scaling problem is the absence of shared semantics. If one system predicts pixels, another predicts latent motion, and a third emits robot commands, their data cannot be compared merely by adopting the same file format. A useful standard must declare the intervention, predicted or intended change, time horizon, reference frame, uncertainty, embodiment assumptions, and observed outcome. These fields provide a common decision context while allowing the underlying encoding to differ.

Existing resources make both the opportunity and the limitation visible. Open X-Embodiment aggregates cross-robot data, while DROID and BridgeData V2 broaden real-world manipulation coverage \citep{oneill2023openx,khazatsky2024droid,bridgedatav2}. UMI supports portable collection of human demonstrations, Ego4D provides egocentric human video, and RLDS defines recording semantics for sequential-decision data \citep{chi2024umi,grauman2021ego4d,rlds2021}. These resources expose different subsets of embodiment, sensing, timing, task, action, and outcome information. None already implements the contract proposed here. The missing fields concern how a model request was represented, translated, verified, executed, and judged.

Evaluation requires the same role-aware structure. RoboWM-Bench tests whether generated behavior can be converted into executable robot actions rather than scoring visual plausibility alone \citep{robowmbench2026}. RoboBench evaluates embodied reasoning, whereas A2Eval automates benchmark curation and evaluation through collaborative agents \citep{luo2025robobench,zhang2026a2eval}. LIBERO-Safety and ROBOSHACKLES examine safety failures in VLA or embodied-foundation-model settings \citep{cui2026liberosafety,yin2026roboshackles}. These efforts answer different questions. The roadmap therefore separates predictive quality, interface grounding, control, tool composition, safety, traceability, and learning progress.

\begin{figure}[tp]
    \centering
    \includegraphics[width=0.88\linewidth]{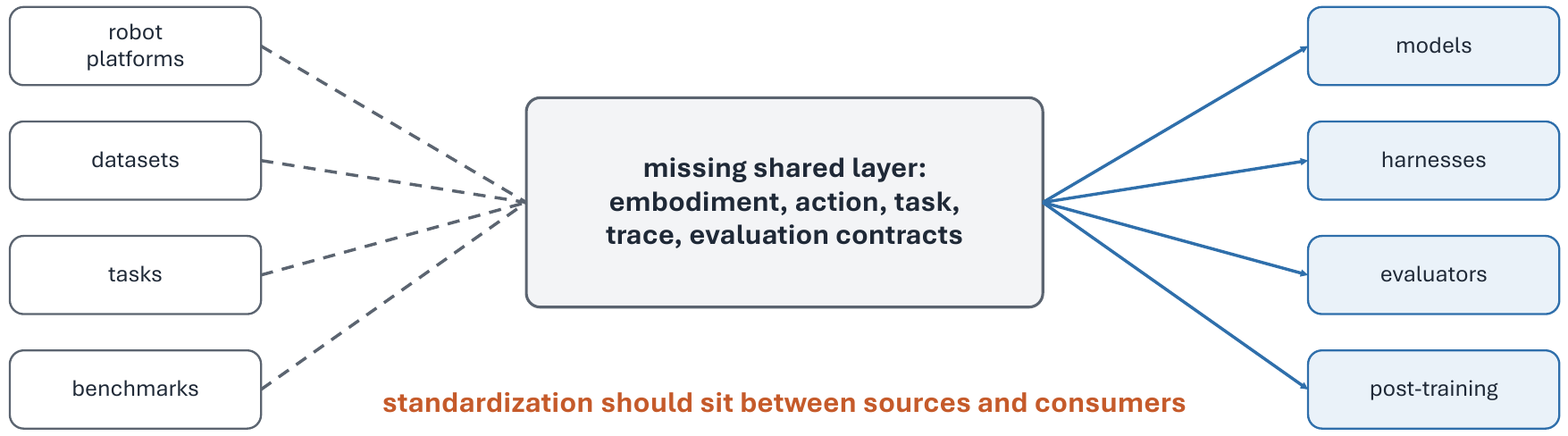}
\caption{Compatibility layer between data producers and system consumers. Robot platforms, datasets, tasks, and benchmarks expose shared embodiment, action, task, trace, and evaluation metadata to models, harnesses, evaluators, and post-training systems.}
\label{fig:missing-standardization-layer}
\end{figure}

Figure~\ref{fig:missing-standardization-layer} depicts the proposed compatibility layer between data-producing artifacts and their consumers. Its arrows represent declared records flowing into shared contracts and those contracts being consumed by models, harnesses, evaluators, and post-training pipelines.

\begin{figure}[tp]
    \centering
    \includegraphics[width=0.90\linewidth]{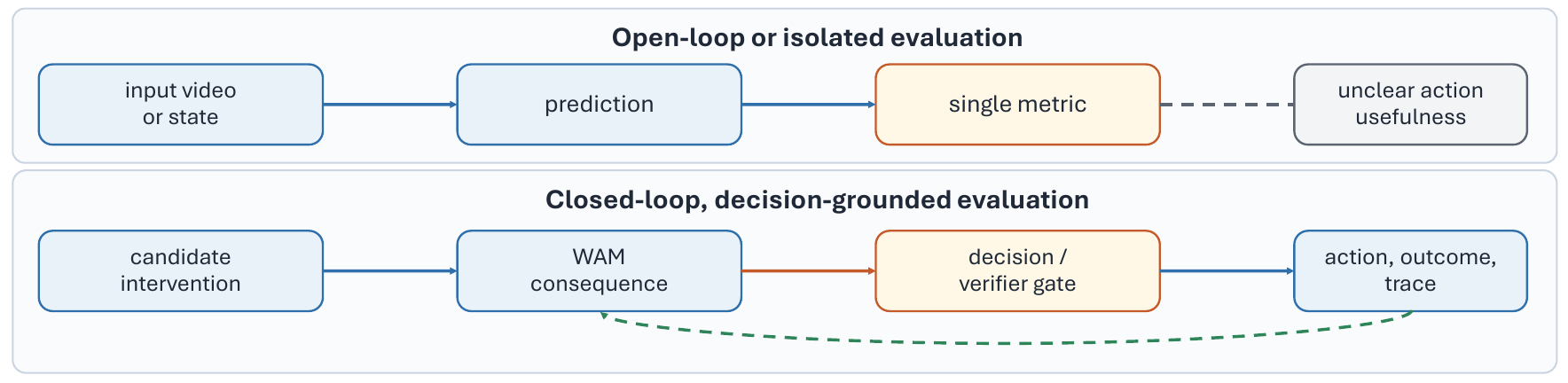}
\caption{Open-loop versus decision-grounded evaluation. A WAM query couples physical context with a candidate intervention, and its consequence declares horizon, frame, uncertainty, and validity. Progress is measured by improved decisions or outcomes, not forecast quality alone.}
\label{fig:open-loop-vs-closed-loop-evaluation}
\end{figure}

Figure~\ref{fig:open-loop-vs-closed-loop-evaluation} contrasts isolated prediction scoring with the decision-grounded evaluation advocated in this paper. The feedback arrow indicates that only outcome-linked traces should return to model assessment or learning.

\subsection{Ecosystem and Systems Gap}

The ecosystem and systems gap concerns how a brain model becomes a physical agent without absorbing the complete agent into one model. General reasoning, low-level stabilization, specialist perception, verification, memory, and execution evolve at different rates. When these functions are hidden inside one project-specific stack, replacing a controller or tool can require retraining and revalidating the entire system. The embodied brain should therefore be distinguished from the broader physical-intelligence stack in which it operates.

Digital agents make this separation visible. Language models operate inside harnesses that manage tool calls, state, guardrails, tracing, approval, and multi-step execution \citep{openai2025agentsguide,openai2025agentssdk,anthropic2024buildingeffectiveagents}. Language, code, and structured APIs provide interfaces that downstream services can parse. Physical agents face a harder communication problem. Learned visual or multimodal tokens may be unusable by a controller unless it shares that representation or has a trained adapter. Geometry, coordinate frames, tactile state, timing, uncertainty, and embodiment constraints must also remain consistent during execution.

A physical harness must consequently do more than call a tool. It must ground the brain-model output, align representations and frames, resolve a declared capability, verify preconditions, synchronize controllers, monitor outcomes, and preserve provenance. Some of these functions may be learned. Others may be implemented through deterministic systems or control software. Their common purpose is to prevent the embodiment-specific execution layer from defining the general reasoning model.

The ecosystem gap extends beyond architecture. A dataset may omit the frame needed to interpret an intended transition. A robot may not declare the controller assumptions required for comparison. A benchmark may discard failures needed for post-training, and a tool may not expose its preconditions or failure modes. Each omission is reasonable within a closed project. Collectively, they prevent independently developed work from contributing to a shared stack.

The required response is not one universal robot, file format, or benchmark. It is a minimum set of contracts that makes heterogeneous systems interpretable. Embodiment profiles, brain-output schemas, coordinate conventions, tool capabilities, harness traces, verifier protocols, task definitions, and evaluation layers serve this role. A longer-term shared 3D/4D representation may reduce translation costs further, but its encoding remains a research question. Standardizing semantics can begin before the field converges on that representation.

Table~\ref{tab:gap-system-map} closes the chapter by mapping each introduction gap to its core question, observable symptoms, and corresponding roadmap response.

\begin{table}[tp]
\centering
\scriptsize
\setlength{\tabcolsep}{2pt}
\renewcommand{\arraystretch}{1.08}
\begin{tabularx}{\textwidth}{>{\columncolor{figgray!5}}P{0.17\textwidth} P{0.29\textwidth} P{0.29\textwidth} Y}
\toprule
\textbf{Barrier} & \textbf{Why progress does not accumulate} & \textbf{Roadmap mechanism} & \textbf{Observable test} \\
\midrule
Model and representation & Prediction and action interfaces expose different variables and often retain local embodiment assumptions. The model role and consequence semantics remain unclear. & Define a WAM prediction contract and an embodied-brain intent interface; leave embodiment-specific execution downstream. & Decision-grounded prediction, grounded intent, tool substitution, and cross-embodiment adaptation. \\
Objective and standardization & Data, tasks, supervision, and metrics describe different decision contexts, so additional trajectories do not automatically become reusable experience. & Standardize interface semantics, Embodiment Cards, Task Cards, Trace Cards, and layered evaluation while preserving raw artifacts. & Cross-dataset replay, reconstructable decisions, comparable layer-specific metrics, and reusable failure traces. \\
Ecosystem and systems & Project-specific runtimes hide representation translation, controller assumptions, verification, and failure handling. Components cannot be replaced independently. & Build a physical harness around declared capability interfaces, provenance, verifier gates, and trace admission. & Controller or tool replacement without full brain retraining, attributable failures, and regression-gated updates. \\
\bottomrule
\end{tabularx}
\caption{Barriers, corresponding roadmap mechanisms, and observable tests for cumulative progress.}
\label{tab:gap-system-map}
\end{table}

\subsection{From Gaps to Roadmap Requirements}

The three gaps support one conclusion: physical intelligence requires scalable interfaces as well as scalable models. The model gap asks what predictive capability should become reusable physical reasoning. The standardization gap asks how heterogeneous tasks and experience can supervise that capability. The systems gap asks how model-level intent becomes executable without binding the model to one controller or body. These requirements constrain one another. The model output determines what must be translated and logged, while the task protocol determines which supervision is available. The tool contract determines which intentions are executable, and evaluation determines which traces can return to learning.

The Roadmap therefore proceeds from a system contract rather than a list of independent agendas. It first defines the roles and interfaces of the complete physical-intelligence stack. It then develops the embodied brain and WAM prediction contract, the physical harness and tool ecosystem, data and task records, closed-loop learning, and community standards. Each stage answers one barrier while producing information needed by the next. The intended payoff is cumulative progress: model reasoning can transfer, embodiment-specific adaptation can remain local, and verified interaction can become reusable experience.

%% file: chapters/04_coevolution_roadmap.tex
\section[Roadmap: Co-evolving Brain Models, Harnesses, Data, and Ecosystems]{Roadmap: Co-evolving Brain Models, Harnesses, Data, and\\Ecosystems}

The central claim of this roadmap is that physical intelligence cannot scale cumulatively while model-level reasoning, embodiment-specific control, data conventions, and runtime logic remain entangled. A larger model may improve one component without making its output reusable. A larger dataset may add trajectories without making their decision contexts comparable. A stronger controller may improve one platform without clarifying what the reasoning model contributed. The proposed response is to co-develop capabilities and the interfaces that connect them.

This is not a claim that modularity or standardization guarantees general intelligence. The roadmap instead defines bounded, testable responsibilities. It asks what the central model should learn, what its output means, how that output becomes executable, what an interaction record must preserve, and which evidence permits an update. These interfaces allow each hypothesis to be evaluated without prescribing one final neural architecture or robot platform.

\subsection{Roadmap at a Glance: System Roles, Interfaces, and Scaling Logic}

The complete system follows one operational flow. Multimodal observations, task context, memory, and available capabilities inform the embodied brain. The brain emits an intermediate intent representation. A physical harness grounds that intent, resolves tool models and controllers, and supervises execution. The resulting world change is recorded as a trace, evaluated, and admitted to learning only when its provenance and quality are sufficient.

\begin{figure}[tp]
    \centering
    \includegraphics[width=\linewidth]{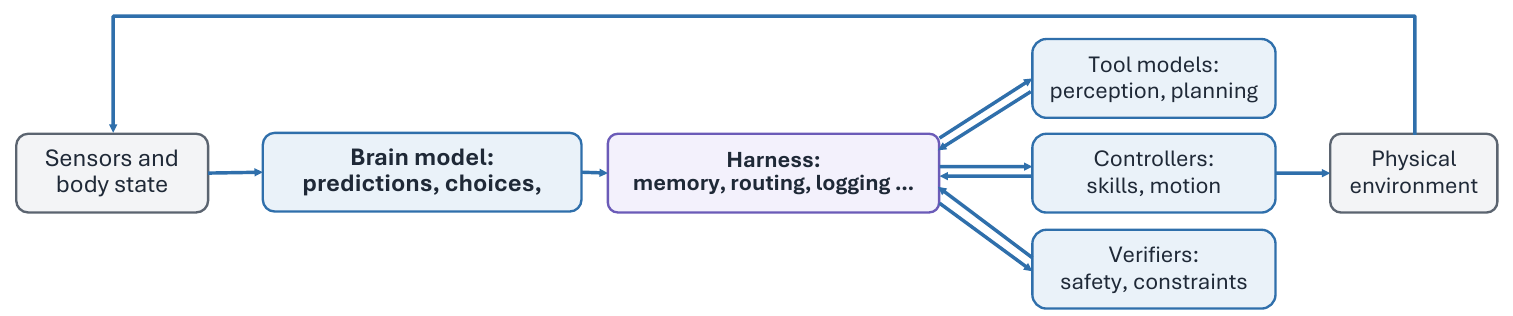}
\caption{Functional boundary between the embodied brain and the surrounding physical-intelligence stack. Sensors and body state inform model predictions and decisions; the harness manages memory, routing, logging, verification, and capability calls; tool models, controllers, and verifiers mediate interaction with the physical environment.}
\label{fig:embodied-brain-system}
\end{figure}

Three distinctions organize this flow. The embodied brain is a model-level target, whereas the physical-intelligence stack is the complete agent system. A WAM is a present research prototype for intervention-conditioned prediction, not a mandatory system layer. A tool model provides specialized perception, prediction, planning, or control, while a tool is the physical or digital capability that is ultimately invoked. Figure~\ref{fig:embodied-brain-system} visualizes these ownership boundaries.

Table~\ref{tab:ownership-boundary} states the same architecture as an inspectable system contract. The separation is functional rather than necessarily architectural: components may be trained jointly when their inputs, outputs, and transformations remain observable.

\begin{table}[tp]
\centering
\scriptsize
\setlength{\tabcolsep}{2.0pt}
\renewcommand{\arraystretch}{1.10}
\begin{tabularx}{\textwidth}{>{\columncolor{figgray!5}}P{0.14\textwidth} P{0.23\textwidth} P{0.38\textwidth} Y}
\toprule
\textbf{Element} & \textbf{Owns} & \textbf{Interface: receives $\rightarrow$ exposes} & \textbf{Explicitly does not own} \\
\midrule
Embodied brain & Evidence integration, intervention comparison, and intended physical change. & Observations, task, memory, embodiment metadata, capability descriptions $\rightarrow$ intended transition or capability request with entities, frames, constraints, uncertainty, and completion or recovery conditions. & Actuator commands, high-frequency stabilization, and hardware-specific control. \\
Physical harness & Intent grounding, capability resolution, verification, execution coordination, and recovery. & Brain intent, runtime state, capability contracts, embodiment constraints, verifier policies $\rightarrow$ tool/controller calls, execution state, monitoring events, recovery decisions, and Trace Cards. & Specialist inference and low-level control. \\
Tool models and controllers & Specialized perception, prediction, planning, simulation, grasping, navigation, control, or verification. & Typed requests, observations or state, frames, constraints $\rightarrow$ predictions, plans or actions, execution status, timing, and failure signals. & Environment response and learning admission. \\
Tools, body, and environment & Sensing, actuation, and physical or digital capability endpoints. & Controller/tool calls and environment conditions $\rightarrow$ observations, acknowledgements, state changes, and outcomes. & Outcome evaluation and model updating. \\
Experience and evaluation loop & Decision reconstruction, evidence evaluation, and trace admission. & Embodiment, Task, and Trace Cards; metrics; verifier, regression, and safety results $\rightarrow$ evaluations, reusable traces, and update-gate status. & Parameter updates and capability training. \\
\bottomrule
\end{tabularx}
\caption{Functional ownership and interfaces in the physical-intelligence stack. Each boundary states what information is received, what is exposed, and what remains downstream.}
\label{tab:ownership-boundary}
\end{table}

Two model interfaces should not be conflated. The proposed \emph{WAM prediction contract} takes physical context and a candidate intervention, then exposes a decision-relevant consequence. It also declares the prediction horizon, reference frame, uncertainty, and validity conditions. The \emph{brain intent interface} instead communicates what change should be realized or which capability should be invoked. A WAM prototype may inform this decision, but its predicted consequence is not itself the final tool or actuator command.

Representation standardization should proceed on two horizons. In the near term, systems can agree on semantics, frames, timing, uncertainty, and capability fields while allowing language, structured state, learned tokens, video, geometry, and hybrid encodings to compete. Over the longer term, a shared world-centric 3D/4D representation may reduce translation across perception, prediction, planning, and control. This convergence is a research objective rather than an established universal solution.

\begin{figure}[tp]
\centering
\resizebox{\textwidth}{!}{
\begin{tikzpicture}[
    every node/.style={font=\sffamily\scriptsize},
    stage/.style={figmodel, minimum width=2.45cm, minimum height=0.95cm, very thick},
    detail/.style={figcontract, minimum width=2.45cm, minimum height=0.58cm},
    num/.style={circle, draw=figblue!85!black, fill=figblue!12, inner sep=1.5pt, font=\sffamily\bfseries\scriptsize}
]
\node[stage] (m) at (0,0) {Brain model};
\node[stage] (h) at (2.95,0) {Tool models\\+ harness};
\node[stage] (d) at (5.90,0) {Data\\+ tasks};
\node[stage] (l) at (8.85,0) {Closed-loop\\learning};
\node[stage] (e) at (11.80,0) {Ecosystem\\coordination};
\node[num] at (-1.12,0.42) {1};
\node[num] at (1.83,0.42) {2};
\node[num] at (4.78,0.42) {3};
\node[num] at (7.73,0.42) {4};
\node[num] at (10.68,0.42) {5};
\node[detail] (md) at (0,-1.00) {predict, decide, query};
\node[detail] (hd) at (2.95,-1.00) {route, execute, verify};
\node[detail] (dd) at (5.90,-1.00) {record, compare};
\node[detail] (ld) at (8.85,-1.00) {learn, test, improve};
\node[detail] (ed) at (11.80,-1.00) {share, align, scale};
\foreach \a/\b in {m/h,h/d,d/l,l/e} \draw[figarrow] (\a) -- (\b);
\foreach \a/\b in {m/md,h/hd,d/dd,l/ld,e/ed} \draw[figcompat] (\a) -- (\b);
\node[figdata, minimum width=2.45cm, minimum height=0.54cm] (trace) at (8.85,-2.05) {admitted physical traces};
\draw[figlearn] (ld.south) -- (trace.north);
\draw[figlearn] (trace.west) .. controls (6.95,-2.65) and (1.1,-2.65) .. (md.south);
\draw[figlearn] (trace.west) .. controls (7.15,-2.35) and (3.2,-2.35) .. (hd.south);
\draw[figlearn] (trace.west) .. controls (7.35,-2.10) and (5.9,-2.10) .. (dd.south);
\node[figtitle, text=figgreen!80!black] at (4.45,-2.92) {feedback updates model, harness, and task/evaluation design};
\end{tikzpicture}}
\caption{Five-stage co-evolution path from embodied-brain capabilities to ecosystem coordination. Solid blue arrows indicate design dependencies, gray lines group each stage with its responsibility, and dashed green arrows return admitted traces to model, harness, and task/evaluation design.}
\label{fig:roadmap-overview}
\end{figure}
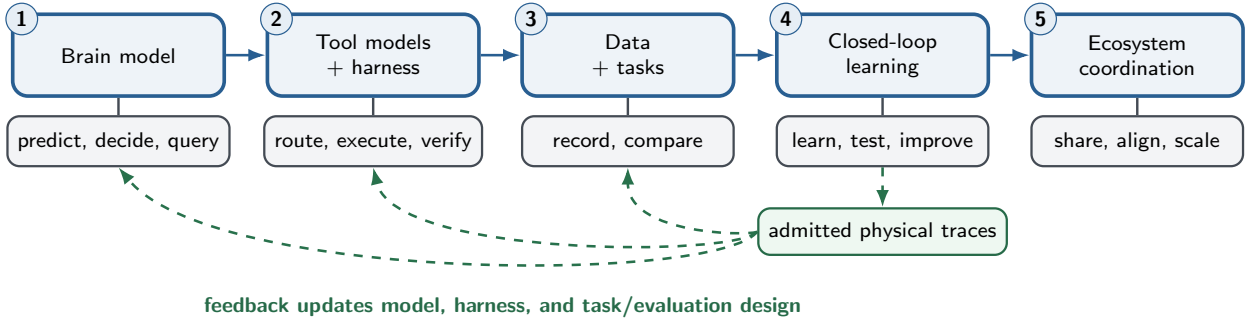

The five stages in Figure~\ref{fig:roadmap-overview} order the development agenda. The solid path indicates design dependencies, not a rigid training pipeline. Admitted traces return evidence to the model, harness, and task design so that the complete stack can improve coherently.

\subsection{Embodied Brain and the WAM Prediction Contract}

A physical-intelligence system should not be reduced to a model that predicts every possible future and directly controls every actuator. The embodied brain is the general reasoning model within the system. It consumes multimodal observations, task context, retrieved memory, embodiment information, and descriptions of available capabilities. Its role is to integrate this evidence, compare possible interventions, and determine an intended physical change. Persistent storage, high-frequency stabilization, and hardware-specific execution can remain outside the model.

The defining output is an intermediate intent representation rather than an actuator trajectory. It should identify an intended state transition or capability request. Depending on the task, it may also include relevant entities or regions, spatial and temporal frames, constraints, uncertainty, and completion or recovery conditions. The encoding remains open. Language or code may be sufficient for some capabilities; structured geometry, learned tokens, or hybrid representations may be required for others. The requirement is that the harness can interpret and validate the output without binding the brain to one gripper, controller, or morphology.

WAM research makes the predictive part of this target experimentally accessible. The proposed prediction contract conditions on physical context and a candidate intervention. It returns a decision-relevant consequence with declared scope and uncertainty. Future observations, latent transitions, geometric states, contact changes, affordances, or task progress can satisfy this role when they improve decision making. RGB video is therefore optional, and its physical content must be evaluated rather than assumed.

An action may be predicted jointly with the consequence, but an action vector alone does not satisfy this contract. The action states what a particular embodiment should execute; the consequence states what the system expects that intervention to change. Keeping both semantics visible allows a model to be evaluated for foresight even when the downstream controller changes.

This distinction preserves the value of present WAMs without fixing the future architecture. WAM prototypes offer practical routes to learning intervention-conditioned prediction from video, robot trajectories, simulation, and interaction. A future embodied brain may absorb these capabilities, reorganize them, or replace the current WAM abstraction. The roadmap specifies what should be learned and communicated while leaving that architectural choice open.

Role separation also does not prohibit joint or end-to-end training. A single network may implement several functions, and gradients may cross the declared boundaries. The requirement is an inspectable contract: the intended change, representation translation, selected capability, executed action, and observed outcome should remain attributable. This makes generalization claims testable rather than assuming that modularity alone produces transfer.
\subsection{Tool Models and Harness: From Decisions to Executable Capabilities}

A harness is central because physical intelligence is not produced by a model alone. Digital agent systems connect language models to tools, state, guardrails, tracing, and multi-step workflows \citep{openai2025agentsguide,openai2025agentssdk,anthropic2024buildingeffectiveagents}. Language, code, and structured API calls make these capabilities addressable. A physical harness inherits this orchestration role, but it must additionally ground an intent in space, time, embodiment, and dynamics before execution.

We distinguish a \emph{tool} from a \emph{tool model}. A tool is a physical or digital capability endpoint, such as a gripper, mobile base, camera, display, browser, or external implement. A tool model is a learned or algorithmic module that supports a capability, such as a grasp planner, navigation policy, state estimator, force controller, simulator, or verifier. One tool model may support several hardware variants through adapters. Conversely, one tool may require several models and controllers. The harness should expose these combinations through a capability registry rather than hard-code them into the brain.

\begin{figure}[tp]
    \centering
    \includegraphics[width=0.5\linewidth]{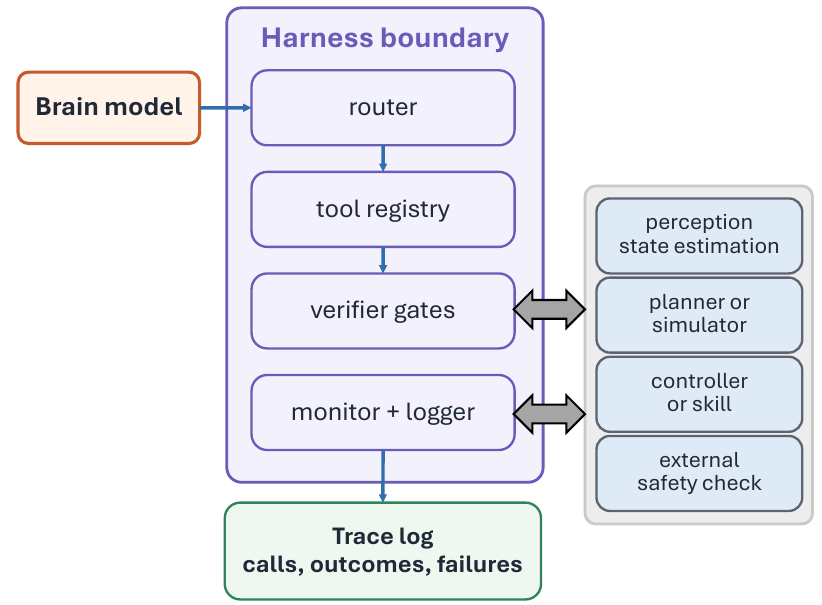}
\caption{Runtime organization of the brain model, harness, and external capabilities. The harness routes requests through a tool registry and verifier gates, invokes perception, planning, control, or safety modules, and records calls, outcomes, and failures in a trace log.}
\label{fig:brain-harness-tool}
\end{figure}

Figure~\ref{fig:brain-harness-tool} summarizes this runtime organization and the trace path used to record capability calls and outcomes.

A capability contract declares accepted inputs, coordinate conventions, preconditions, controllable variables, outputs, timing, and failure modes. This contract allows tools to be composed hierarchically. For example, a request to drive a peg can be resolved into arm motion, gripper control, hammer acquisition, and a constrained impact. The same brain may also request a digital capability, such as opening an interface or displaying information. In both cases, the brain specifies an intended effect and the harness constructs the executable chain.

Communication is the hardest part of this separation. A learned visual token has no operational meaning for a controller unless the controller shares its representation or receives an appropriate adapter. The interface must preserve entities, coordinate frames, geometry, timing, uncertainty, and relevant physical constraints. Near-term schemas can make these semantics explicit even when their encodings differ. Longer-term research can investigate whether a shared world-centric 3D/4D representation reduces the need for pairwise adapters.

Uni-Inter provides a bounded indication of this direction. Its Unified Interactive Volume represents human--human, human--object, and human--scene interactions in a common volumetric field \citep{liu2025uniinter}. This result does not establish a universal brain--harness language. It shows, in a narrower setting, that heterogeneous interaction entities can be aligned within one spatial representation.

The physical harness may therefore include learned adapters as well as conventional software. It can translate brain outputs into world-centric state, align frames, maintain multimodal runtime memory, resolve capabilities, synchronize control, verify preconditions, monitor execution, and record provenance. A limited functional analogy is the nervous system, which mediates between central reasoning and heterogeneous sensory or motor structures. The analogy motivates an adaptive communication layer; it does not prescribe a biological architecture.

\begin{figure}[tp]
\centering
\includegraphics[width=0.96\linewidth]{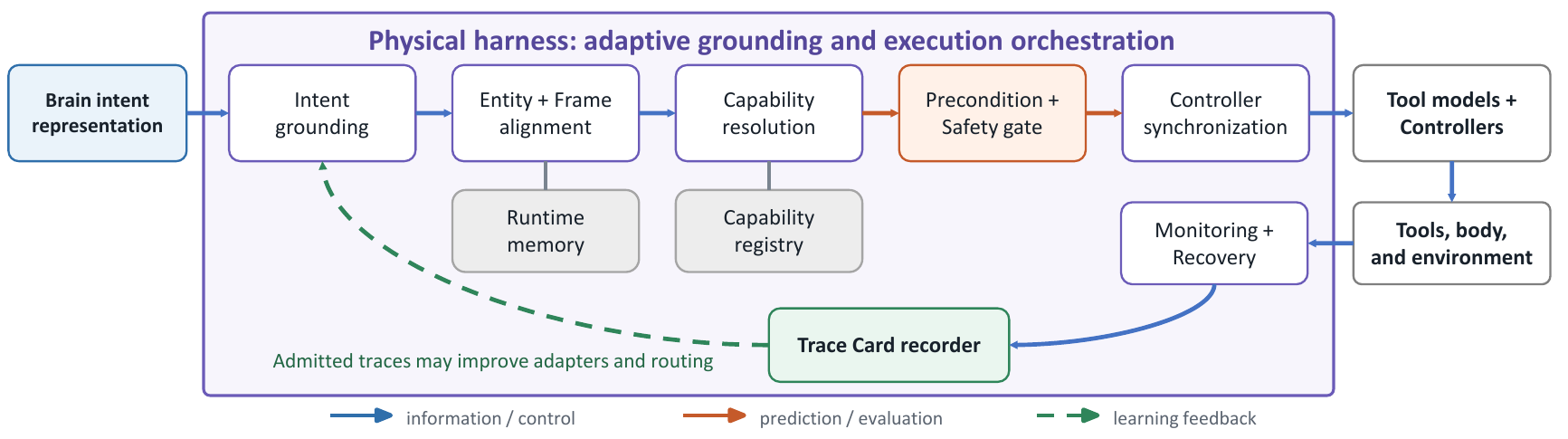}
\caption{Internal responsibilities of the physical harness. It grounds the brain intent, aligns entities and frames, resolves capabilities, checks preconditions, synchronizes controllers, monitors execution, and records a Trace Card. Runtime memory, the capability registry, and admitted traces support adaptation and routing.}
\label{fig:physical-harness-internals}
\end{figure}

Figure~\ref{fig:physical-harness-internals} details how these responsibilities connect intent grounding to monitored execution and trace recording.

This design creates concrete tests. A controller should be replaceable without relearning the complete brain. An adapter should be evaluated through tool and embodiment substitution. A verifier should reject infeasible requests without excessive false vetoes. Memory, tracing, and recovery should improve failure attribution and replay. The harness is valuable when it makes model-level reasoning executable, inspectable, and reusable across such changes.

\begin{figure}[tp]
    \centering
    \includegraphics[width=0.82\linewidth]{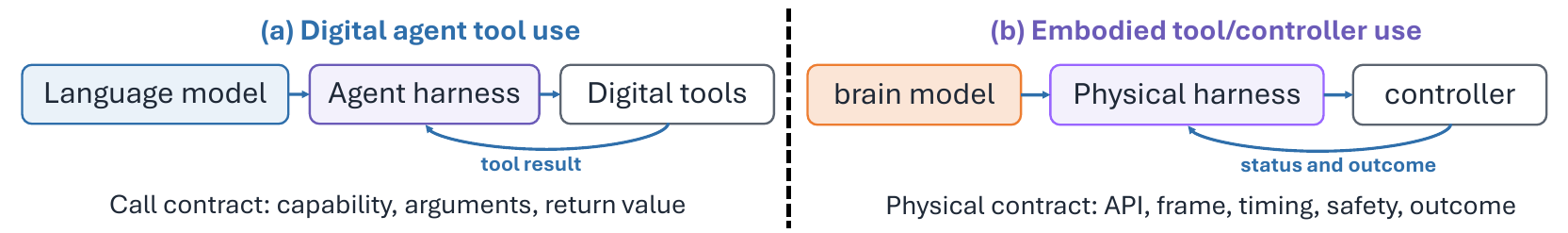}
\caption{Analogy between digital and physical tool use. A language model can address tools through natural language, code, or structured APIs, whereas an embodied brain relies on a physical harness to ground intent, resolve capabilities, enforce timing and safety constraints, and coordinate controllers.}
\label{fig:tool-calling-in-robotics}
\end{figure}

Figure~\ref{fig:tool-calling-in-robotics} contrasts this physical execution path with digital tool calling. Both require orchestration, but physical interaction additionally requires grounding, temporal coordination, and embodiment-aware constraints.

\begin{figure}[tp]
\centering
\includegraphics[width=0.96\linewidth]{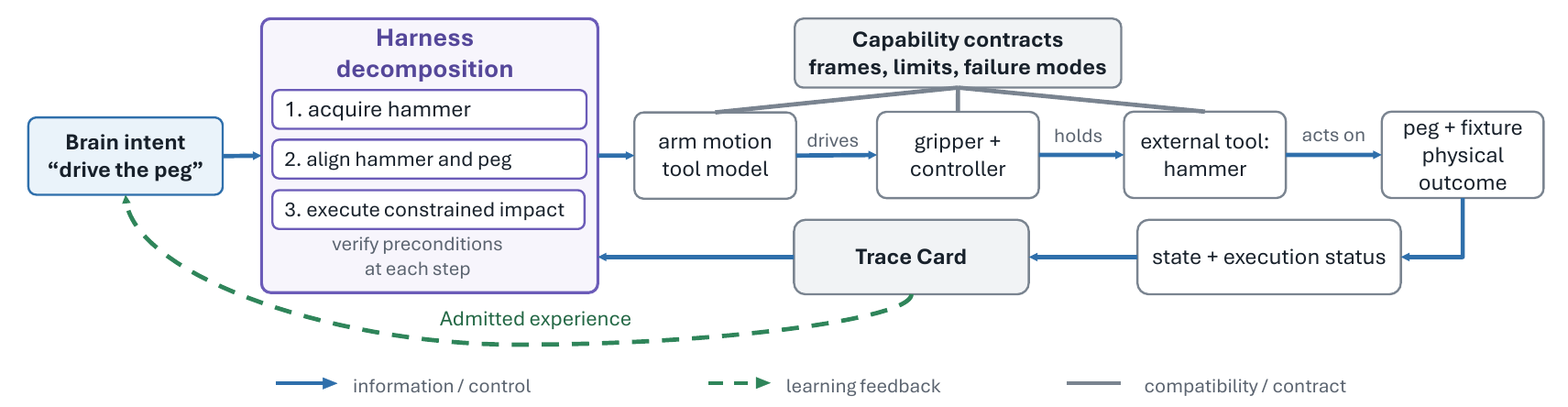}
\caption{Hierarchical composition of physical capabilities for a constrained impact task. The harness decomposes one intent into arm, gripper, and external-tool calls. Capability contracts expose frames, limits, and failure modes; execution status returns through a Trace Card for diagnosis and learning.}
\label{fig:hierarchical-tool-composition}
\end{figure}

Figure~\ref{fig:hierarchical-tool-composition} illustrates hierarchical composition through a constrained impact task. The example is intentionally decomposed: arm motion drives a gripper, the gripper holds an external implement, and the implement acts on the environment. This makes each capability and failure boundary independently describable.

\FloatBarrier

\subsection{Data Construction and Task Building: Contracts for Learning and Comparison}

Data construction should follow the interfaces that models and systems need to learn. Once a WAM prediction declares its intervention, horizon, frame, and consequence, a dataset can record whether that consequence occurred. Once a brain output declares an intended transition or capability request, a trace can record how the harness interpreted and executed it. This alignment turns data collection from signal accumulation into the construction of reusable decision contexts.

The proposal does not require one robot, sensor suite, storage format, or control interface. It requires enough metadata to distinguish shared semantics from local implementation. A manipulation trace without calibration, reference frames, action bounds, controller assumptions, or failure annotations may support imitation learning. It provides much weaker evidence for consequence modeling, interface transfer, or tool substitution.

We propose three lightweight records. An \emph{Embodiment Card} describes the body, sensors, coordinate conventions, calibration, control rates, action bounds, available tools, and safety limits. A \emph{Task Card} describes the goal, scene context, permitted observations and capabilities, constraints, success criteria, and meaningful failure modes. A \emph{Trace Card} links synchronized observations to the brain request, representation version, selected tool chain, verifier decisions, controller state, execution outcome, corrections, and data-quality flags.

These cards define a replayable decision context rather than a new mandatory file format. Robot, human, simulation, and harness data can retain their original artifacts while mapping relevant fields into the contract. This preserves embodiment-specific information without allowing those details to remain hidden.

Task construction should follow the same principle. A benchmark should specify which physical variables matter, which observations and tools are available, and which failures should be diagnosed. It should also state which intermediate predictions or interface transformations are evaluated. Data and tasks then share one semantic contract: the training record contains the evidence that the evaluation protocol later requires.

Table~\ref{tab:data-task-contract} summarizes the minimum fields and scaling role of the three proposed records. The fields are deliberately semantic so that existing datasets and runtimes can map to them without adopting one storage format.

\begin{table}[tp]
\centering
\scriptsize
\setlength{\tabcolsep}{2pt}
\renewcommand{\arraystretch}{1.08}
\begin{tabularx}{\textwidth}{>{\columncolor{figgray!5}}P{0.18\textwidth} P{0.35\textwidth} Y}
\toprule
\textbf{Record} & \textbf{Minimum fields} & \textbf{Scaling role} \\
\midrule
Embodiment Card & Morphology, sensors, frame conventions, calibration, tools, action bounds, control rates, safety limits, and controller assumptions. & Makes observations, intended effects, and executed actions interpretable across robot platforms and simulators. \\
Task Card & Goal, scene, object state, allowed observations, allowed tools, safety constraints, success criteria, and meaningful failure labels. & Turns tasks into reusable training and evaluation settings rather than isolated demonstrations. \\
Trace Card & Observations, brain request, representation schema and frame, candidate interventions, selected tool chain, adapters, predictive outputs, verifier outputs, controller states, outcomes, corrections, and data-quality flags. & Supports brain--harness interface learning, predictive modeling, tool-chain supervision, safety evaluation, post-training, and regression tests. \\
\bottomrule
\end{tabularx}
\caption{Minimum records for reusable physical experience, preserving shared decision semantics together with embodiment-specific execution details.}
\label{tab:data-task-contract}
\end{table}

\subsection{Closed-Loop Physical Learning: Model, Harness, and Environment in the Loop}

The roadmap should move beyond static pretraining and one-shot imitation. Language-model development offers two useful patterns. InstructGPT illustrates staged supervised and preference-based post-training \citep{ouyang2022training}. Agent frameworks place a model inside a runtime that manages tools, state, tracing, guardrails, and evaluation \citep{openai2025agentsguide,openai2025agentssdk,anthropic2024buildingeffectiveagents}. TeleBoost provides a separate temporal-generation example of decomposing expensive alignment into explicit optimization stages and stability gates \citep{liang2026teleboost}. These systems are implementation references, not evidence that their recipes transfer directly to robotics.

Physical interaction introduces harder feedback. Outcomes are partially observed, depend on the embodiment, cost more to collect, and can be unsafe. The model may produce a valid intent even when the controller fails, or an invalid intent may appear successful under a forgiving task. Learning therefore requires a trace that separates model reasoning, representation translation, capability selection, control, and outcome.

The appropriate unit is harness-in-the-loop physical learning. The brain proposes an intended transition or capability request. The harness grounds it, invokes predictive or verification models when needed, resolves a tool chain, and executes an approved action. The resulting trace records both the decision and its realization. An evaluator then determines whether the evidence is suitable for representation learning, predictive training, interface alignment, verifier learning, reinforcement learning, distillation, or regression testing.

This loop connects to broader proposals for learning from experience and autonomous machine intelligence \citep{silver2025eraexperience,lecun2022path}. RoboCat, Voyager, Eureka, and DreamerV3 offer distinct examples of self-generated data, skill accumulation, reward design, or world-model learning \citep{bousmalis2023robocat,wang2023voyager,ma2023eureka,hafner2023dreamerv3}. Their domains and assumptions differ. They motivate mechanisms to test rather than establishing that unconstrained physical self-improvement is solved.

Evaluation is the admission gate that prevents a self-reinforcing error loop. It should test whether predictions improve decisions, brain outputs remain grounded, tools can be substituted, failures are recoverable, and updates preserve prior capability and safety. These properties should be reported separately instead of compressed into one end-task score.

Table~\ref{tab:learning-stages} separates the proposed learning objectives by signal, updated component, and required gate. The rows are complementary objectives rather than a mandatory chronological pipeline.

\begin{table}[tp]
\centering
\scriptsize
\setlength{\tabcolsep}{2pt}
\renewcommand{\arraystretch}{1.06}
\begin{tabularx}{\textwidth}{>{\columncolor{figgray!5}}P{0.17\textwidth} P{0.24\textwidth} P{0.22\textwidth} Y}
\toprule
\textbf{Learning objective} & \textbf{Primary signal} & \textbf{Updated component} & \textbf{Required gate} \\
\midrule
Physical representation learning & Images, videos, robot observations, depth, tactile, force, simulation, and egocentric data. & Encoders, state estimators, and world-centric representations. & Geometry, contact, affordance, temporal consistency, and cross-view alignment. \\
Predictive physical reasoning & Future observations or states, latent transitions, inverse dynamics, counterfactuals, progress, and optional video. & Current WAM prototypes or corresponding capabilities within a brain model. & Decision-grounded ablation, physical consistency, uncertainty calibration, and counterfactual consistency. \\
Brain--harness interface alignment & Brain requests paired with frames, adapters, tool capabilities, execution states, and outcomes. & Output representation, interface adapters, and capability resolver. & Representation grounding, adapter dependence, tool substitution, and cross-embodiment transfer. \\
Tool-chain supervision & Goals, intended transitions, tool calls, verifier calls, recovery choices, refusals, and approval decisions. & Brain deliberation and harness routing policies. & Tool preconditions, call correctness, compositional execution, and recovery quality. \\
Verifier alignment & Unsafe futures, vetoes, near misses, human corrections, and preference comparisons. & Verifier, safety head, and decision policy. & False vetoes, missed hazards, and safety retention. \\
Harness-in-the-loop RL & Rewards, outcomes, failed attempts, simulation replay, and real interaction traces. & Brain model, interface adapters, harness policies, tool models, verifiers, and capability registry. & Safety filters, trace quality, regression tests, and rollback criteria. \\
Distillation and adaptation & Expert traces, slow reasoning plans, high-frequency control logs, and platform logs. & Smaller policies, runtime adapters, and controllers. & Latency, timing stability, recovery behavior, and cross-embodiment transfer. \\
\bottomrule
\end{tabularx}
\caption{Physical post-training objectives, their supervision, updated components, and admission gates. The rows are complementary rather than a rigid chronology.}
\label{tab:learning-stages}
\end{table}

Table~\ref{tab:layered-evaluation} complements these learning objectives with a role-aware evaluation protocol. It separates predictive reasoning, interface grounding, control, orchestration, safety, traceability, and update quality so that an end-task score does not conceal the source of improvement or failure.

\begin{table}[tp]
\centering
\scriptsize
\setlength{\tabcolsep}{1.8pt}
\renewcommand{\arraystretch}{1.08}
\begin{tabularx}{\textwidth}{>{\columncolor{figgray!5}}P{0.15\textwidth} P{0.25\textwidth} P{0.09\textwidth} Y}
\toprule
\textbf{Layer} & \textbf{Question} & \textbf{Status} & \textbf{Representative measurements} \\
\midrule
Predictive physical reasoning & Does prediction capture decision-relevant physical change? & \bothbadge & Visual fidelity, physical consistency, action plausibility, and embodied executability \citep{wang2026wam,robowmbench2026}; proposed transition/contact accuracy, counterfactual consistency, and uncertainty calibration. \\
Brain--harness grounding & Is intent grounded and preserved? & \proposedbadge & Entity and frame consistency, adapter dependence, capability resolution, tool substitution, and cross-embodiment transfer. \\
Policy and control & Can the declared embodiment execute the action? & \existingbadge & Task success, action smoothness, controller/contact feasibility, latency, and cross-embodiment adaptation. \\
Tool orchestration & Are tools selected and composed correctly? & \proposedbadge & Call precision, parameter validity, precondition satisfaction, compositional execution, and tool-failure recovery. \\
Verification and safety & Are risky or infeasible plans rejected? & \bothbadge & Physical, semantic, and injury-related hazards \citep{cui2026liberosafety,yin2026roboshackles}; proposed false-veto, missed-hazard, reversibility, and unsafe-action rejection measures. \\
Harness and traceability & Are failures attributable and traces reusable? & \proposedbadge & Trace completeness, failure attribution, provenance, replay determinism, and regression coverage. \\
Self-improvement & Do updates improve capability without reducing safety? & \proposedbadge & Learning progress, failure recurrence, safety retention, regression tests, and rollback triggers. \\
\bottomrule
\end{tabularx}
\caption{Layered evaluation of physical intelligence. Status distinguishes measurements already used in the cited literature from prospective roadmap extensions; ``Both'' combines the two.}
\label{tab:layered-evaluation}
\end{table}

\subsection{Ecosystem Roadmap: Shared Standards for Cumulative Progress}

The final part of the roadmap concerns cumulative community progress. Physical intelligence will remain expensive to scale if each model, robot, dataset, benchmark, and tool hides a different interface. Shared agreements can make these artifacts compatible without imposing one hardware design or learning method. Their purpose is not uniformity. Their purpose is to make specialization reusable.

The first agreements should be operational. A robot declares its embodiment and controller assumptions. A brain model declares the semantics of its output. A harness declares its representation adapters and routing decisions. A tool declares its capabilities, preconditions, and failure modes. A dataset records the decision context, and a benchmark states which layer it evaluates. Embodiment, Task, and Trace Cards provide a lightweight starting point for these declarations.

Such contracts change how work accumulates. A new controller can be evaluated beneath an existing intent interface. A new predictive model can be tested inside the same decision protocol. A dataset can support future objectives because its frames, interventions, and outcomes remain interpretable. A benchmark can localize improvement instead of reporting only an end-to-end score. The ecosystem succeeds when independent contributions can be compared and combined without erasing their local assumptions.

\subsection{Near-Term Research Agenda: From WAM Prototypes to an Embodied Brain}

The roadmap does not treat a WAM as a solved module waiting to be inserted into an embodied brain. It treats WAM research as an experimental route toward predictive and interventional capabilities that a future brain model may require. Three near-term priorities make this route measurable.

The first priority is a decision-relevant WAM prediction contract. Current systems span future-state and action prediction, video--action modeling, 3D world modeling, and cascaded or joint predictive-control designs~\citep{guo2024predictionactionvisualpolicy,cheang2024gr2generativevideolanguageactionmodel,li2025unifiedvideoactionmodel,lu2025gwm,wang2026wam}. Video prediction can provide dense temporal targets and inspectable futures~\citep{finn2017deepvisualforesightplanning,ebert2018visualforesightmodelbaseddeep,zhou2024robodreamerlearningcompositionalworld,liang2022simple}. Its usefulness should be tested through geometry, contact, uncertainty, and decision improvement rather than inferred from visual quality~\citep{robowmbench2026}. A concrete milestone is a benchmark in which models declare horizon, frame, consequence semantics, and confidence, then demonstrate that these predictions improve intervention selection or verification.

The second priority is a grounded brain--harness representation. Near-term work should compare structured state, language, learned tokens, 3D/4D representations, and hybrid interfaces under one semantic contract. The relevant tests are whether the harness recovers the intended entities, frames, constraints, and state changes. Stronger evidence comes from substituting a tool or embodiment while retaining the brain-level request. Adapter cost and residual performance loss should be reported rather than hidden inside end-to-end fine-tuning.

The third priority is an adaptive physical harness and trace ecosystem. Capability registries, representation adapters, multimodal memory, verifier integration, monitoring, and recovery can each be evaluated before a general embodied brain exists. The harness should support hierarchical tool composition across controllers, body parts, external implements, and digital services. A concrete milestone is a replayable Trace Card that attributes failure to intent, translation, tool selection, control, or environment response. Candidate updates should then pass regression and safety gates before promotion.

Together, these priorities define an achievable transition from WAM prototypes to an embodied brain and its surrounding physical-intelligence stack. They localize embodiment-specific adaptation while preserving a shared reasoning objective. They also convert data collection, interface design, execution, and evaluation into mutually informative stages. The long-term destination remains ambitious, but consequence contracts, grounded representations, substitutable tools, replayable traces, and gated updates can be built and tested now.

\FloatBarrier

%% file: chapters/05_conclusion.tex
\section{Conclusion}

This paper has surveyed the overlapping trajectories from action-centric policies, language-conditioned robot learning, and predictive world models toward WAMs. Their common opportunity is intervention-conditioned physical reasoning: estimating what may change under an action and using that estimate to improve a decision. Current WAMs provide promising prototypes for this capability. They do not determine the final architecture of a general physical agent.

The central limitation is not model capacity alone. Action outputs remain meaningful only relative to an embodiment and controller. Predictive interfaces expose different consequence variables, while data, tasks, and runtimes encode different assumptions. These dependencies restrict reuse even when individual systems improve. We described them through model and representation, objective and standardization, and ecosystem and systems gaps.

The co-evolution roadmap addresses these gaps through explicit responsibilities. A WAM prediction contract exposes a decision-relevant consequence with its scope and uncertainty. The embodied brain serves as the long-term model target that reasons over context and interventions, then communicates an intended state transition or capability request. A physical harness grounds that intent through declared tool models, controllers, and verifiers. Embodiment, Task, and Trace Cards preserve the information required to reconstruct and evaluate the resulting interaction.

This organization changes how physical intelligence can scale. General physical reasoning can be improved without treating one actuator space as its permanent output. Embodiment-specific adaptation can remain visible in tools and adapters. Compatible traces can support comparison, failure attribution, and regression-gated post-training. These benefits remain hypotheses to test through decision-grounded prediction, representation grounding, tool substitution, cross-embodiment adaptation, and safe learning progress.

The roadmap does not claim that interfaces alone produce AGI or that one representation should be fixed in advance. It provides a technically actionable path by which models, data, tools, and systems can improve separately while contributing to a shared physical-intelligence stack. If these interfaces mature, heterogeneous physical experience can become cumulative rather than project-specific. That shift may bring adaptive and self-improving agents closer to general intelligence grounded in sustained interaction with the physical world.